\documentclass[prx,superscriptaddress,aps,amsmath,amssymb,floatfix,reprint,raggedbottom]{revtex4-2}
\usepackage[margin=1in]{geometry}
\usepackage{balance}
\usepackage{hyperref}

\hypersetup{
    colorlinks=true,
    linkcolor=blue,
    citecolor=blue,
    urlcolor=blue
}

\usepackage{graphicx}

\usepackage{url}

\usepackage{multirow}
\usepackage{siunitx}
\usepackage{threeparttable}
\usepackage{makecell}
\usepackage{array}
\usepackage{adjustbox}  
\usepackage[linesnumbered,ruled,vlined]{algorithm2e}

\DeclareUnicodeCharacter{2009}{\,}

\makeatletter
\renewcommand{\fnum@figure}{\textbf{Fig.~\thefigure}}
\makeatother

\AtBeginDocument{%
\makeatletter
\long\def\@makecaption#1#2{%
  \vskip\abovecaptionskip
  {\footnotesize #1.
#2\par}%
  \vskip\belowcaptionskip}%
\makeatother
}

\makeatletter
\def\bbordermatrix#1{\begingroup \m@th
  \@tempdima 4.75\p@
  \setbox\z@\vbox{%
    \def\cr{\crcr\noalign{\kern2\p@\global\let\cr\endline}}%
    \ialign{$##$\hfil\kern2\p@\kern\@tempdima&\thinspace\hfil$##$\hfil
      &&\quad\hfil$##$\hfil\crcr
      \omit\strut\hfil\crcr\noalign{\kern-\baselineskip}%
      #1\crcr\omit\strut\cr}}%
  \setbox\tw@\vbox{\unvcopy\z@\global\setbox\@ne\lastbox}%
  \setbox\tw@\hbox{\unhbox\@ne\unskip\global\setbox\@ne\lastbox}%
  \setbox\tw@\hbox{$\kern\wd\@ne\kern-\@tempdima\left[\kern-\wd\@ne
    \global\setbox\@ne\vbox{\box\@ne\kern2\p@}%
    \vcenter{\kern-\ht\@ne\unvbox\z@\kern-\baselineskip}\,\right]$}%
  \null\;\vbox{\kern\ht\@ne\box\tw@}\endgroup}
\makeatother

\emergencystretch=\maxdimen
\usepackage[compact]{titlesec}
\titlespacing{\section}{0pt}{*3}{*2}
\titlespacing{\subsection}{0pt}{*2}{*2}
\titlespacing{\subsubsection}{0pt}{*2}{*2}

\titleformat{\section}{\normalfont\small \bfseries}{\thesection.}{1em}{}

\usepackage{booktabs}

\begin{document}

\title{IsingFormer: Augmenting Parallel Tempering With Learned Proposals}


\author{Saleh Bunaiyan}
\affiliation{Department of Electrical \& Computer Engineering, University of California, Santa Barbara, Santa Barbara, CA 93106, USA}

\affiliation{Electrical Engineering Department, King Fahd University of Petroleum \& Minerals,
Dhahran 31261, Saudi Arabia}

\author{Corentin Delacour}
\affiliation{Department of Electrical \& Computer Engineering, University of California, Santa Barbara, Santa Barbara, CA 93106, USA}

\author{Shuvro Chowdhury}
\affiliation{Department of Electrical \& Computer Engineering, University of California, Santa Barbara, Santa Barbara, CA 93106, USA}

\author{Kyle Lee}
\affiliation{Department of Electrical \& Computer Engineering, University of California, Santa Barbara, Santa Barbara, CA 93106, USA}

\author{Abdelrahman S. Abdelrahman}
\affiliation{Department of Electrical \& Computer Engineering, University of California, Santa Barbara, Santa Barbara, CA 93106, USA}

\author{Kerem Y. Camsari}
\affiliation{Department of Electrical \& Computer Engineering, University of California, Santa Barbara, Santa Barbara, CA 93106, USA}

\begin{abstract}
Generative models have been extensively used to accelerate MCMC mixing for sampling and optimization, but their effective integration with standard MCMC remains an open question. Here, we introduce a global proposal move in which finite-temperature configurations from an external generator are used as proposals within Parallel Tempering (PT). We examine a specific generator, IsingFormer, a Transformer trained on long-run MCMC configurations intended to approximate equilibrium distributions, and call the resulting framework Transformer-Augmented Parallel Tempering (TAPT). The IsingFormer exhibits two useful capabilities: interpolation to untrained $\beta$ values and conditional completion under clamped settings absent from training. On 3D spin-glass instances, TAPT reaches substantially lower residual energies than standard PT in fewer Monte Carlo sweeps. On integer factorization, we show how a structured problem encoding can amortize the training cost by training IsingFormer once and reusing the same model across target products that were not imposed during training. Finally, in a scaling study that uses long-run MCMC configurations as proposals, with proposal generation excluded from the timing, TAPT reduces the fitted time-to-solution exponent by approximately $33\%$ relative to PT over the tested problem sizes.
\end{abstract}
\maketitle

Generative models are powerful at proposing structured candidates, but are often weak verifiers. Recent work has coupled generative models with standard MCMC methods to accelerate ground-state search~\citep{delbono2026demonstrating}. This is similar in spirit to the concept of \emph{generator}--\emph{verifier} collaborations, particularly in reasoning tasks, where a generator proposes candidates at scale and a separate rule-based verifier checks them. For example, in theorem proving, the generator is often a Transformer and the verifier is a proof checker~\cite{trinh2024solving}. The verifier provides guarantees that the generator alone cannot.

In this work, we use energy evaluation as the exact candidate-quality check and Monte Carlo dynamics as the search mechanism. Standard Markov chain Monte Carlo (MCMC) can preserve a known target distribution. Parallel Tempering (PT), also known as replica-exchange Monte Carlo \cite{Hukushima1996JPSJ}, strengthens MCMC with nonlocal swaps across a temperature ladder while retaining local moves within each replica. Despite the nonlocal nature of PT, solving optimization or sampling problems in rugged landscapes remains difficult due to the long mixing and equilibration times. 

We propose to augment PT with finite-temperature configurations used as global proposals. Among the possible ways of generating these configurations, we examine a specific generative model, \emph{\textbf{IsingFormer}}, a Transformer \citep{vaswani2017attention} trained on long-run MCMC configurations to produce full-system proposals conditioned on the inverse temperature, $\beta$. Inspired by methods such as Boltzmann Generators \cite{doi:10.1126/science.aaw1147}, IsingFormer provides \emph{independent} nonlocal proposals that capture global structure. Within the proposed framework, these proposals are accepted or rejected using a $\Delta E$-only Metropolis criterion, after which PT continues with local Gibbs updates and replica exchanges. This $\Delta E$-only proposal rule does not preserve the Boltzmann target in general and is used here as an optimization heuristic. For unbiased equilibrium sampling, the full Metropolis--Hastings correction involving proposal probabilities is required~\cite{hastings1970monte}. IsingFormer learns global structure that accelerates ground-state search within standard Monte Carlo methods. It does not solve each optimization instance on its own.

\begin{figure*}[t!]
    \centering
    \includegraphics[width=\linewidth]{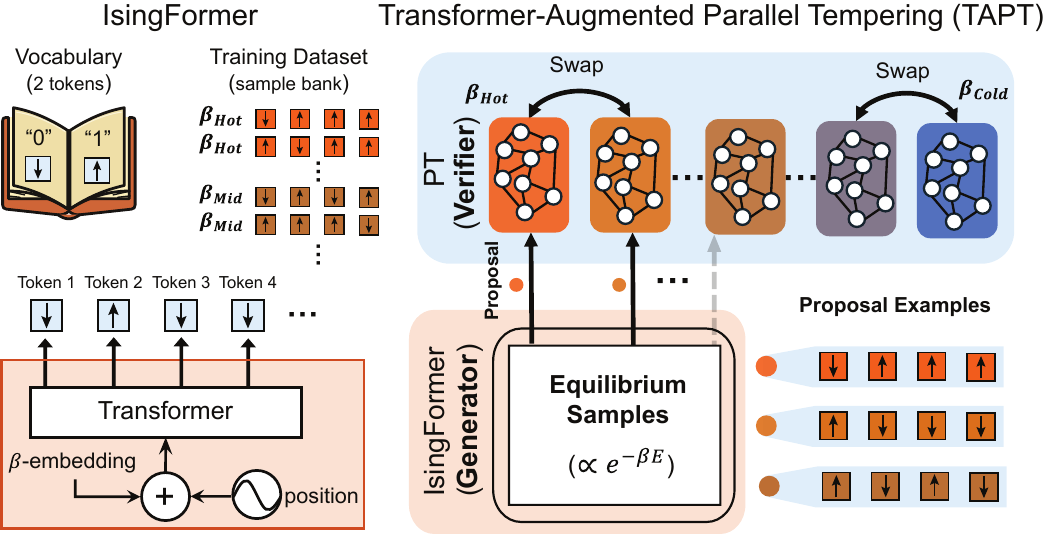}
    \caption{\textbf{IsingFormer and the Transformer-Augmented Parallel Tempering (TAPT) framework.} \textbf{Left:} Configurations generated by long-run MCMC at several inverse temperatures $\beta$, intended to approximate the corresponding equilibrium distributions, form the IsingFormer training sample bank. Binary spins are represented using a two-token vocabulary. IsingFormer, a decoder-only Transformer, combines positional information with a learned $\beta$ embedding and autoregressively models $q_{\theta}(x\mid\beta)$, enabling the generation of complete spin configurations at selected temperatures. \textbf{Right:} These configurations are introduced as nonlocal proposals to hot and intermediate replicas of the PT ladder. Proposal energies are evaluated, proposals are filtered using the optimization-biased $\Delta E$-only acceptance rule, and local MCMC updates and neighboring-replica swaps allow accepted low-energy states to propagate toward the coldest replicas. Because representative training data at high-$\beta$ (low temperature) are computationally costly to generate, the coldest replicas are not augmented with learned proposals.  
}
    \label{fig:Figure1}
\end{figure*}

We call the resulting algorithm \textbf{Transformer-Augmented Parallel Tempering (TAPT)}. TAPT keeps PT's inverse-temperature ($\beta$) ladder and replica-exchange dynamics, but interleaves them with global proposal moves (Fig.~\ref{fig:Figure1}). This provides a general mechanism for incorporating finite-temperature proposal generators into PT while retaining its local Monte Carlo and replica-exchange dynamics. We do not augment the coldest replicas, since for NP-hard optimization instances, obtaining representative training data at very high $\beta$ can itself amount to solving the original problem.

We demonstrate the utility of this approach in four settings.
(1) On the 2D Ising model, IsingFormer reproduces the free-energy and magnetization curves and exhibits two important capabilities: interpolation to unseen $\beta$ values, including the critical region, and conditional completion under unseen clamped settings.
(2) On 3D spin-glass instances, TAPT lowers the residual energy much faster than standard PT, with a rapid drop when proposal moves are activated.
(3) For integer factorization, a structured problem encoding allows the training cost to be amortized across instances. A single IsingFormer trained on the free-running multiplier circuit is reused across target products that were not imposed during training by changing only the clamped product bits at inference.
(4) In a scaling study for integer factorization with long-run MCMC proposals in place of IsingFormer, TAPT reduces the fitted TTS scaling exponent by approximately $33\%$ relative to PT.

Prior work on accelerating sampling or optimization problems with neural networks can be broadly categorized into two approaches: \textbf{(1)} training models to learn the full equilibrium (Boltzmann) distribution for sampling, and \textbf{(2)} training models to directly find low-energy states for optimization. Our work combines the two: a model trained to approximate equilibrium sampling serves as the proposal generator inside a Monte Carlo optimizer.

\textbf{Learned Boltzmann and Ising Samplers.} There is a large body of work that learns equilibrium distributions to bypass long MCMC mixing times \citep{jung2025roadmap}. Self-learning Monte Carlo introduced learned global updates with an exact acceptance correction~\citep{liu2017selflearning}. Early Boltzmann-machine studies learned finite-temperature Ising distributions from Monte Carlo data and generated configurations reproducing thermodynamic observables, including near criticality \citep{torlai2016learning,morningstar2018deep}. Variational autoregressive networks (VANs) fit the Boltzmann law and provide  uncorrelated samples and free-energy estimates \citep{wu2019solving, ma2024message}. RBMs and autoregressive models have been used as Metropolis–Hastings proposal generators  to accelerate MCMC \citep{Huang2017PRB, Wang2017PRE, wu2021unbiased, PhysRevE.101.023304}. Generative proposals have also been used for equilibrium spin-glass Monte Carlo~\citep{scriva2023accelerating}. Diffusion generators have  been benchmarked on 2D Ising models \citep{lee2025thermodynamic,bae2025diffusion}. Discrete diffusion models based on tensor networks have further been integrated with MCMC to generate nonlocal proposals for equilibrium sampling in spin systems
\citep{causer2025discrete}. Replica exchange between stacked RBMs has been shown to improve mixing time for MNIST, lattice proteins, and 2D Ising models \citep{ICLR2024_34e278fb}.

Beyond Ising models, Boltzmann Generators have been applied to molecular equilibrium sampling and free-energy estimation \citep{doi:10.1126/science.aaw1147}. Normalizing-flow-based replica exchange and autoregressive samplers have further been developed for molecular and materials systems \citep{invernizzi2022skipping,damewood2022sampling}. In lattice settings, normalizing flows serve as Metropolis-Hastings proposal generators \citep{PhysRevD.107.014512,PhysRevD.100.034515,PhysRevLett.125.121601,boyda2021sampling,PhysRevD.106.074506}.

Learned generative models have also been integrated directly with Parallel Tempering in continuous-state settings. Temperature-steerable flows have been used to provide MCMC proposals across multiple temperatures while retaining conventional replica exchanges \citep{dibak2022temperature}, while a recent neural-transport extension of Parallel Tempering modifies the replica-communication mechanism using learned transport maps \citep{zhang2026accelerated}. Related PT extensions include variational reference distributions~\citep{surjanovic2022parallel}.

We adopt a learned finite-temperature sampler as the generator but integrate it into PT for discrete Ising configuration spaces with an explicit $\beta$ conditioning. The independent proposals are accepted with a $\Delta E$-only criterion, allowing low-energy proposals to propagate down the inverse-temperature ($\beta$) ladder during optimization.

\textbf{Learning for Combinatorial Optimization.} Another body of work uses neural models with the primary goal of finding low-energy or ground-state configurations for optimization. Examples include variational neural annealing and related annealing or tempering schemes that use learned generative models to access low-energy and ground-state configurations in Ising and quadratic unconstrained binary optimization (QUBO) problems
\citep{Hibat_Allah_2021,ma2024message,mcnaughton2020boosting,zhang2025generative}. Related work in quantum many-body physics uses Transformers as neural wavefunction ans\"atze for variational ground-state searches \citep{sprague2024variational}.

Diffusion models have been developed as conditional solvers for graph optimization problems such as maximum independent set (MIS), MaxCut, MaxClique, and minimum dominating set (MDS), generalizing across unseen instances within a problem class \citep{sanokowski2024diffusion,sanokowskiscalable}. Related energy-based graph-learning approaches train real-valued extensions of Ising Hamiltonians so that low-energy states encode desired outputs, with inference performed by energy minimization \citep{ICLR2024_fcc4c8ca}.

Machine-learning-assisted annealing has also been used for spin-glass optimization, combining learned global proposals with local Monte Carlo moves under a sequential temperature schedule \citep{delbono2026demonstrating}. Physics-inspired GNNs instead optimize relaxed QUBO Hamiltonians directly \citep{schuetz2022combinatorial}, while reinforcement learning has been used to learn temperature-control policies for simulated annealing that generalize across spin-glass instances \citep{mills2020finding}.

The closest optimization approaches to TAPT use learned equilibrium models as global Monte Carlo proposals during sequential or global annealing \citep{mcnaughton2020boosting,delbono2026demonstrating}. In these methods, the temperature is progressively lowered and the neural model is retrained or transferred along the annealing trajectory. In contrast, TAPT retains the simultaneous replica ladder and replica-exchange dynamics of Parallel Tempering while using a fixed generator to provide additional global proposals. For optimization, TAPT does not require evaluating the generator likelihood or enforcing detailed balance, allowing proposals to be generated and stored offline before the PT run and separating the generator from the online Monte Carlo search.

\begin{figure*}[t!]
    \centering
    \includegraphics[width=\linewidth, keepaspectratio]{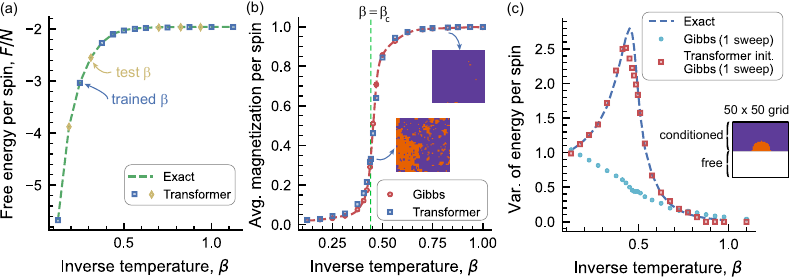}
    \caption{\textbf{Equilibrium learning on 2D Ising (50$\times$50, open-boundary conditions):} (a) Free energy per spin $f(\beta) = F(\beta)/N$ from Kac-Ward determinant formalism (exact, green dashed) vs. transformer-generated samples (yellow/blue). Blue squares mark trained $\beta$ values, while yellow diamonds mark test $\beta$ values. The model reproduces $f(\beta)$ at trained points and interpolates to unseen $\beta$. (b) Absolute magnetization vs. $\beta$, showing the finite-size crossover near the critical point $\beta_c \approx 0.44$. Transformer samples (blue) match Gibbs sampling (red); insets show representative configurations obtained from the transformer. (c) Energy variance per spin obtained from the numerical second derivative of $-\beta F$ (blue dashed) vs. single-sweep estimates: Gibbs (cyan) and transformer-initialized Gibbs after one full sweep (red), markedly improving agreement. Each data point is averaged over 100 runs. Insets show the half-clamped setup, unseen during training. The exact free energy (see Supplementary \ref{app:sec_context_exp}) was re-computed for this clamped condition before evaluating the variance.}
    \label{fig:Figure2}
\end{figure*}

\section*{RESULTS}

Before coupling a generator to PT, we must show the generator can accurately learn equilibrium physics. We therefore evaluate IsingFormer on the 2D Ising model where thermodynamics are known exactly. Our evaluation has three parts that we will use later: \textbf{(1)} Unconditioned sample quality via free energy comparisons between IsingFormer and the exact solution, \textbf{(2)} \emph{temperature generalization} across unseen $\beta$ and \textbf{(3)} conditional completion under clamped settings. These establish IsingFormer as a high-quality proposal generator for TAPT. 

The 2D  ferromagnetic Ising model is  a stringent benchmark because its equilibrium statistics are known exactly via the Onsager solution at the thermodynamic limit \cite{Onsager1944}, and finite-size free energies can be computed exactly using the Kac-Ward determinant formalism \cite{Kac1952} and related Pfaffian methods. This enables a direct, quantitative comparison between model-generated samples and ground truth. 

We trained a decoder-only transformer (IsingFormer) on configurations generated by long-run MCMC for a $50 \times 50$ Ising grid with open boundary conditions at several inverse temperatures $\beta$ (see Methods). These configurations are intended to approximate the corresponding equilibrium distributions. The objective is for the learned distribution $q_{\theta}(x\mid\beta)$ to match the Boltzmann law $\pi_\beta(x)$, so that Transformer samples reproduce the correct equilibrium observables. Critically, no free-energy terms are used at training time.

Fig.~\ref{fig:Figure2}(a) shows that the transformer reproduces the exact free energy per spin, $f(\beta)=F(\beta)/N$, not only at the trained temperatures but also at unseen  values of $\beta$. The ability to interpolate in $\beta$ indicates that the model has learned nontrivial structure of the Boltzmann distribution rather than overfitting to individual training distributions. Interpolation is key for our TAPT algorithm, as it allows a single trained model to provide high-quality proposals for replicas across a range of temperatures while simplifying the training.  

Beyond free energy, the transformer also captures ordering behavior. Fig.~\ref{fig:Figure2}(b) shows that the mean absolute magnetization per spin obtained from transformer samples follows the expected sigmoidal curve, including the finite-size crossover near the critical point $\beta_c \approx 0.44$. The ability to reproduce the sharp change in magnetization is especially significant because correlations become long-ranged near criticality, making this region difficult for purely local samplers. Inset configurations confirm that the generated states are physically consistent across ordered and disordered regimes, suggesting that the transformer is able to encode long-range correlations characteristic of the critical point.

The importance of transformer proposals becomes most evident in constrained settings. Fig.~\ref{fig:Figure2}(c) considers the energy variance per spin, a quantity sensitive to sampling errors, under a half-clamped boundary condition. Pure Gibbs sampling from random initialization with a single sweep produces poor estimates, but transformer-initialized Gibbs sampling after just one sweep, i.e. one update of all spins, aligns closely with the exact result obtained from the second derivative of $-\beta F$. This result demonstrates that the transformer can act as a conditional sampler, placing the system much closer to the correct constrained structure. The subsequent Gibbs sweep is a local relaxation step. One sweep alone does not guarantee equilibration. Supplementary Fig.~\ref{fig:collage_clamped} shows this effect: while long Gibbs runs ($10^4$ sweeps) are required to reconstruct the correct boundary-induced structure, a single transformer sample followed by one Gibbs sweep already yields qualitatively similar configurations.

\begin{algorithm}[t]
\SetKwInOut{Input}{Input}
\SetKwInOut{Output}{Output}

\caption{Transformer-Augmented Parallel Tempering (TAPT)}
\label{alg:augPT}

\Input{
Ising energy function $E$, number of replicas $N_{\mathrm{replica}}$,
inverse temperatures $\{\beta_r\}$, number of Monte Carlo sweeps
$N_{\mathrm{MCS}}$, number of sweeps between swap attempts $M$,
and the set of proposal-enabled replicas
$\mathcal{R}_{\mathrm{prop}}$
}
\Output{final replica configurations $\{S_r\}$ and energies $\{E_r\}$}

Initialize replica states $\{S_r\}$ randomly\;

\For{$n \gets 1$ \KwTo $N_{\mathrm{MCS}}/M$}{

    \tcp{Local MCMC sweeps}
    \For{$m \gets 1$ \KwTo $M$}{
        \For{$r \gets 1$ \KwTo $N_{\mathrm{replica}}$}{
            Perform one MCMC sweep on $S_r$ at $\beta_r$\;
        }

    }
Compute $E(S_r)$ for all replicas\;
Store the results in $\{E_r\}$\;
    \tcp{Transformer proposals}
    \ForEach{$r \in \mathcal{R}_{\mathrm{prop}}$}{
        Retrieve the next pre-generated proposal $S_r^{T}$ at $\beta_r$\;
        Compute $E_r^{T} \gets E(S_r^{T})$\;
        $P_{\mathrm{accept}}
        \gets
        \min\!\left[
        1,\exp\!\left(\beta_r(E_r-E_r^{T})\right)
        \right]$\;

        Accept $S_r \gets S_r^{T}$ with probability $P_{\mathrm{accept}}$\;

        \If{proposal is accepted}{
            $E_r \gets E_r^{T}$\;
        }
    }

    \tcp{Replica exchange}
    \If{$n$ is odd}{
        Attempt swaps of states and corresponding energies between
        replica pairs $(1,2),(3,4),\ldots$ with probability
        $P_{\mathrm{swap}}$\;
    }
    \Else{
        Attempt swaps of states and corresponding energies between
        replica pairs $(2,3),(4,5),\ldots$ with probability
        $P_{\mathrm{swap}}$\;
    }
}

\Return{$\{S_r\},\{E_r\}$}\;

\end{algorithm}

\subsection*{Global proposals within parallel tempering}
\label{sec:tapt_opt}

We now augment Parallel Tempering (PT) for optimization with finite-temperature configurations used as global proposals at different temperatures. Standard PT uses replicas of a system at increasing inverse temperatures $\beta_r$, with each replica sampling from its corresponding Boltzmann distribution at $\beta_r$ \cite{Swendsen_1986,Hukushima1996JPSJ}. To improve mixing, PT periodically attempts to swap the states of neighboring replicas with the acceptance probability:
\begin{equation}
    P_{\text{swap}}=\min[1,\exp(\Delta \beta \Delta E)] \label{Pswap}
\end{equation}
with $\Delta \beta=\beta_{r+1}-\beta_r $, and the energy difference between replicas $\Delta E=E_{r+1}-E_r$. Swaps are global moves that enable low-energy samples found at high temperatures to reach colder replicas.

We implement these global moves using IsingFormer, with transformer samples generated at the same PT temperatures to accelerate optimization. Thanks to the transformer's generalization capability across $\beta$ values, it can generate samples at new temperature points within the training range, providing flexibility in the choice of the $\beta$-schedule. For a given replica $r$ at inverse temperature $\beta_r$, the corresponding transformer proposal is accepted with probability:
\begin{equation}
P_{\text{accept}}=\min[1,\exp(\beta_r \Delta E_r)]
    \label{equation: proposal acceptance}
\end{equation}
where $\Delta E_r = E_r - E^T_r$ is the energy difference between replica $r$ and the transformer proposal. Proposals with lower transformer energy are therefore always accepted under the $\Delta E$-only Metropolis criterion. This rule does not preserve the Boltzmann target for the independent proposals considered here, as shown in Supplementary Section~\ref{TAPT derivation}, and we use it solely as an optimization heuristic. For unbiased equilibrium sampling, the independent Metropolis--Hastings move would instead require the appropriate proposal-probability correction, provided the normalized proposal probabilities can be evaluated~\citep{hastings1970monte}.

The Transformer-Augmented Parallel Tempering (TAPT) algorithm, outlined in Algorithm~\ref{alg:augPT}, combines local MCMC updates with two types of global moves: transformer proposals and replica exchanges. The process begins by randomly initializing spin configurations for all replicas. TAPT performs $N_{\mathrm{MCS}}/M$ cycles. In each cycle, every replica undergoes $M$ local MCMC sweeps, followed by proposal moves on the replicas in $\mathcal{R}_{\mathrm{prop}}$ and a replica-exchange stage. For these proposal-enabled replicas (global move 1), proposals are accepted with probability $P_{\text{accept}}$. The coldest replicas are excluded from transformer proposals, as generating representative configurations at low temperatures is computationally costly. Following the transformer proposals, a set of neighboring replica pairs makes swap attempts (global move 2) with probability $P_{\text{swap}}$. At every swap attempt, we alternate between even--odd replica pairs (e.g., $(2,3)$, $(4,5)$) and odd--even replica pairs (e.g., $(1,2)$, $(3,4)$).

\begin{figure*}[t!]
    \centering
    \includegraphics[width=0.98\linewidth, keepaspectratio]{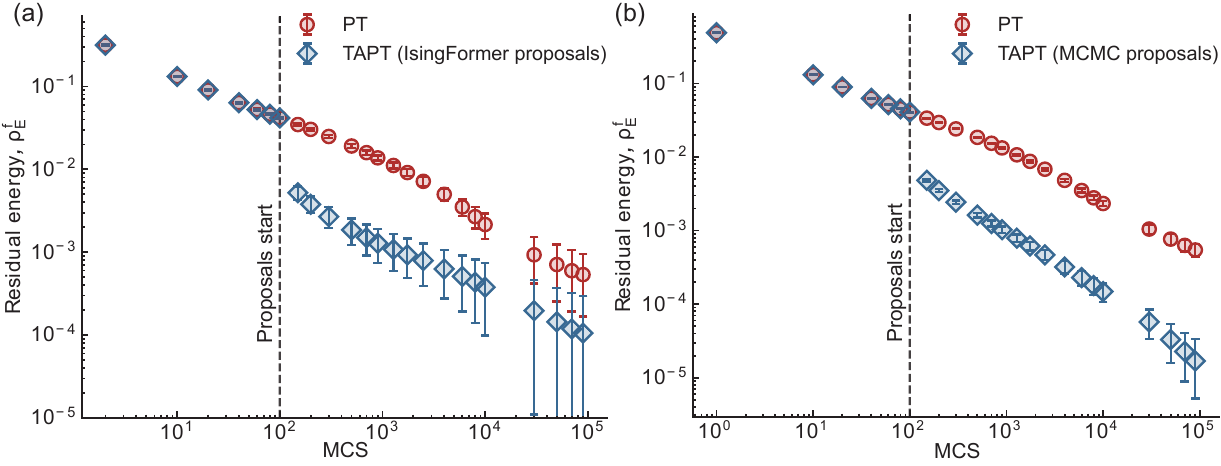}
    \caption{\textbf{TAPT on 3D spin-glass instances (\(L=10\)).} Residual energy $\rho_{E}$ is shown as a function of Monte Carlo sweeps (MCS). For TAPT, proposal moves are enabled after 100 MCS, as indicated by the vertical dashed line, and are subsequently attempted once per MCS. For both PT and TAPT, neighboring-replica swaps alternate between even--odd and odd--even pairings. Proposals are applied to replicas 1--20 of the 22-replica ladder; the two coldest replicas are not proposal-enabled. \textbf{(a)} TAPT using IsingFormer-generated proposals, evaluated on 20 spin-glass instances. \textbf{(b)} TAPT using proposals generated by long-run MCMC, evaluated on 300 spin-glass instances. For each instance, the residual energy is estimated from 400 independent runs. Markers show the mean across instances, and error bars indicate the 95\% bootstrap confidence interval.}
    \label{fig: 3d spin glass}
\end{figure*}

We now evaluate TAPT as an optimizer on two complementary tasks. First, we study 3D spin-glass instances to isolate the effect of global proposals in rugged energy landscapes. Second, we use integer factorization to assess generalization across problem instances that share an underlying structure but differ in their clamped outputs. Details of the IsingFormer training setup and wall-clock time are reported in Supplementary~\ref{supp: IsingFormersTraining}.

\begin{figure*}[t!]
    \centering
    \includegraphics[width=\linewidth, keepaspectratio]{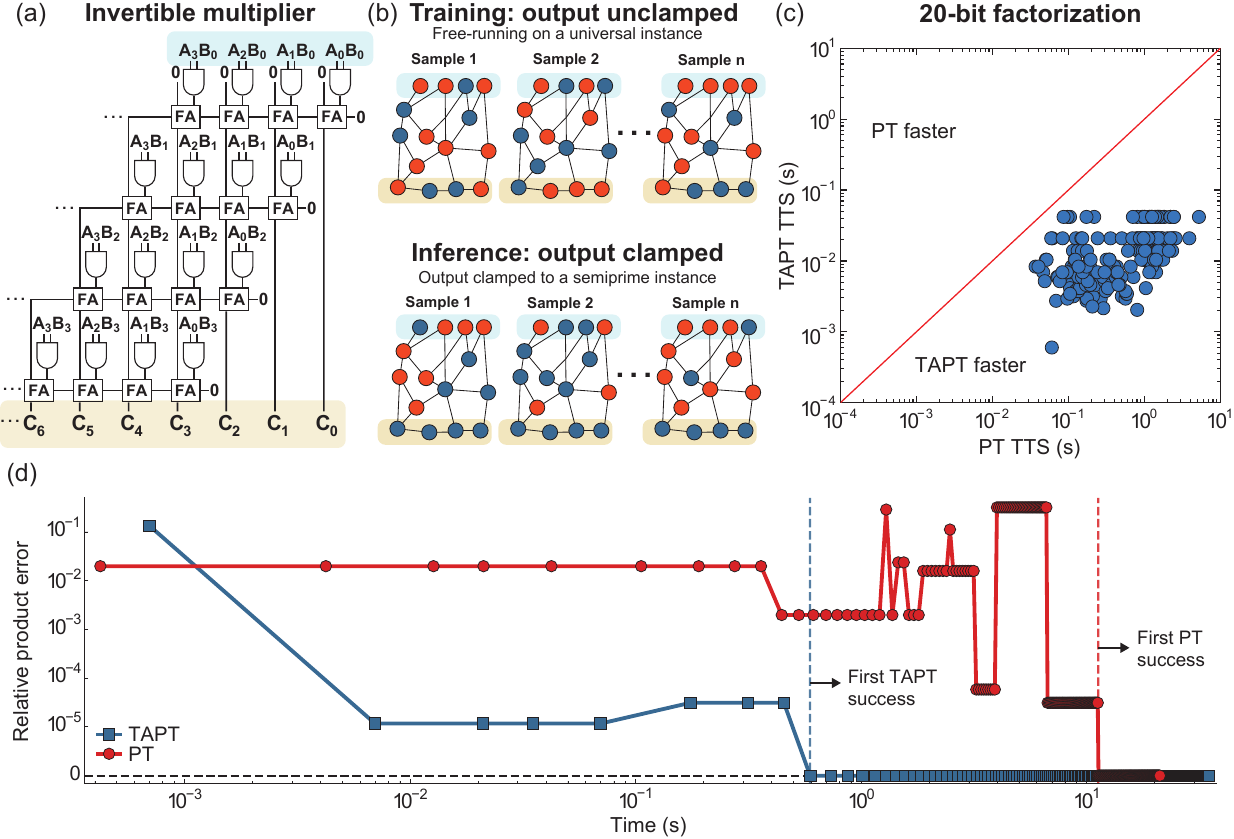}
    \caption{\textbf{TAPT for semiprime factorization.} \textbf{(a)} Integer factorization is formulated as an invertible multiplier circuit encoded by an Ising energy. Operating the circuit in reverse by clamping the output product \(C\) defines an instance-specific optimization landscape whose ground states encode valid factors \(A\) and \(B\). \textbf{(b)} During training, configurations are generated by long-run MCMC from the universal instance with all product spins unclamped, and IsingFormer learns full spin configurations conditioned on inverse temperature. During inference, the output spins are clamped to a target semiprime \(C\), thereby defining a specific factorization instance while reusing the same trained IsingFormer. \textbf{(c)} End-to-end optimization time-to-solution (TTS), conditional on a pre-generated proposal bank, at a target success probability of \(0.99\) for PT and TAPT over 440 semiprime instances. \textbf{(d)} Example run for \(C=1{,}022{,}117=1009\times1013\). The relative product error is defined as \(|AB-C|/C\). The vertical dashed lines indicate the first successful factorization obtained by each method.}
    \label{fig:Figure 4}
\end{figure*}

\subsection*{3D spin-glass optimization} 

We demonstrate TAPT for ground-state search on multiple instances of a 3D spin-glass problem with $L=10$, as shown in Fig.~\ref{fig: 3d spin glass}. We first optimize the temperature ladder using an adaptive scheduler \cite{Isakov2015optimising,chowdhury2025pushing} and then run both PT and TAPT using the same schedule (see Methods). Unless otherwise stated, we use $22$ replicas, $M{=}1$ local sweep per replica between swaps, and a total of $10^5$ Monte Carlo sweeps (MCS), where one sweep corresponds to updating all variables. In TAPT, replicas 1--20 are proposal-enabled, while the two coldest replicas are not augmented.

We report \emph{residual energy} of the replica with the best energy state across the ladder, $\rho_{E} = \langle E - E_{\rm gnd} \rangle / N$, where $N$ is the number of spins and $E_{\rm gnd}$ is the putative ground-state energy. For all instances, we average over $400$ independent runs.

Fig.~\ref{fig: 3d spin glass}(a) shows that TAPT produces a sharp early drop in $\rho_{E}$ at the onset of generator proposals (vertical dashed line), and continues to descend faster than PT as local MCMC updates are interleaved with global proposal and replica-exchange moves. To assess whether this behavior is specific to IsingFormer, we test a variant of TAPT in which the IsingFormer proposals are replaced by long-run MCMC proposals. Figs.~\ref{fig: 3d spin glass}(a) and (b) show that IsingFormer and MCMC proposals produce qualitatively similar behavior. We also study the effect of providing warm-start proposals to PT; however, after the proposals are terminated, PT dynamics alone do not appear to preserve the initial \emph{residual energy} advantage, as shown in Supplementary Fig.~\ref{fig:TAPT_discontinuous_proposal}. In these instances, the proposals are generated \emph{independently} of the current replica state (no-context). Providing partial-state context to the generator did not improve performance in our experiments (Supplementary Fig.~\ref{fig:replica,context}).

The generator supplies nonlocal proposals that escape basins, while energy evaluation filters them and local MCS updates and replica exchange continue the search. Relying on generator proposals alone would rarely reach the ground state, since finite-temperature samples at intermediate $\beta$ seldom land on it. Here, we trained one IsingFormer per spin-glass instance. The generator does not transfer to other instances (its proposals were entirely rejected). Transformer training time is not included in the optimization performance, and without generalization this is a serious limitation shared by many neural optimizers. However, as we discuss next, IsingFormer can generalize to target products not imposed during training when the problem formulation allows different instances to be expressed through simple conditioning. In these cases, model training time can be amortized over inference across many problem instances, accelerating optimization across the family. TAPT is compared here with standard PT, not with enhanced classical baselines such as PT with isoenergetic cluster moves~\citep{Zhu2015PRL,chowdhury2025pushing}.

\subsection*{Generalization across integer-factorization instances}
\label{sec:factorization}

We next evaluate TAPT on a family of Ising instances that share one multiplier structure and differ only in the clamped product: semiprime factorization via an invertible multiplier circuit. Integer factorization can be formulated as an invertible digital circuit, or equivalently as a circuit-SAT problem; see Supplementary \ref{supp: Invertible_Logic_Gates} for the Ising encoding of the multiplier circuit. A logical multiplier $A \times B = C$ can be implemented using invertible AND gates and full adders \cite{andriyash2016boosting,smithson2019efficient,pervaiz2019weighted,aadit2022massively}. Running the circuit in reverse by clamping the output $C$ yields an instance-specific optimization landscape for each semiprime $C$, with the factors $(A,B)$ corresponding to ground states. This setting naturally tests whether learned proposals \emph{generalize across instances}. We use factorization, formulated as a structured inverse-logic problem, as a hard optimization benchmark with an easily verifiable solution, rather than as a competitive factoring algorithm.

In the 3D spin-glass experiment, each instance at a given size is characterized by a distinct energy landscape with its own interaction weights $J_{ij}$ between nodes, hindering the transfer of learned landscapes across instances. As a result, one must account for the \textit{training cost per instance} (see Table~\ref{tab:Trained_IsingFormer}). In contrast, the invertible logic formulation defines one energy model shared across all integer factorization instances at a fixed multiplier size: all instances use the same $J_{ij}$ interactions and differ only through the clamped product bits. Accordingly, we train a single IsingFormer using samples generated from a multiplier circuit with $A,B=10$-bit and $C=20$-bit, where all spins are free-running and no product bits are clamped. During inference, we reuse the same trained IsingFormer and vary only the clamped product bits to specify the target semiprime instance, see Figs.~\ref{fig:Figure 4} (a)--(b). In this setting, the \textit{training cost can be amortized}, since the model is trained once and then reused for inference across many semiprime instances. Similar to the previous IsingFormers, we train on a discrete set of $\beta$ values and interpolate in $\beta$ at inference.

\begin{figure}[t!]
    \centering
    \includegraphics[width=0.98\linewidth, keepaspectratio]{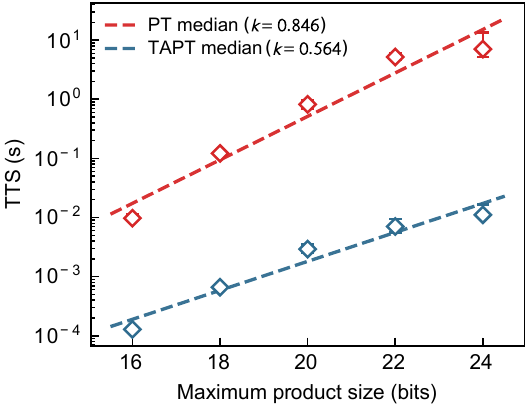}
    \caption{\textbf{End-to-end optimization TTS scaling for semiprime factorization.} The median TTS at a target success probability of \(0.99\), conditional on pre-generated proposal banks, is shown as a function of problem size, quantified by the maximum product bit length. TAPT timing includes all operations performed during optimization, while offline proposal-bank generation is excluded. For this scaling analysis, TAPT uses proposals drawn from long-run MCMC configurations rather than IsingFormer. Markers show the median TTS across instances, and error bars indicate the 95\% bootstrap confidence intervals. The dashed lines are exponential fits of the form \(\mathrm{TTS}(n)=A\exp(kn)\), giving \(k=0.846\) for PT and \(k=0.564\) for TAPT. Equivalently, TAPT reduces the fitted exponent by approximately \(33\%\).}
    \label{fig:Figure TTS}
\end{figure}

For 20-bit factorization, we derive a $\beta$-schedule using the same adaptive procedure as in the 3D spin-glass study and observe that the optimized schedule remains consistent across the family of 20-bit semiprime factorization problems. We use 18 replicas, with proposals applied to the 13 hottest replicas. In TAPT, the output $C$ is provided to the IsingFormer as input tokens, together with the carry-in bits, to condition the generated proposals. We compare PT and TAPT in terms of their time-to-solution (TTS), defined as: 
\begin{equation}
    \mathrm{TTS}
    =
    t_s 
    \left\lceil
    \frac{\log(1-0.99)}
         {\log(1-P_{\mathrm{success}})}
    \right\rceil 
    \label{eq:tts-mcs}
\end{equation}
where $t_s$ is the end-to-end optimization runtime in seconds for a single run of the corresponding duration, conditional on a pre-generated proposal bank, and $P_{\mathrm{success}}$ is the probability of finding a valid solution within that time. For PT, $t_s$ covers the complete optimization run. For TAPT, it includes every operation performed during the run, proposal retrieval and acceptance included, only the offline generation of the proposal bank is excluded (see Methods). Success is defined as satisfying $A\times B=C$, regardless of the internal multiplier configuration. The results in Fig.~\ref{fig:Figure 4}(c) show that TAPT achieves a lower TTS than PT for all 440 tested semiprime instances. Fig.~\ref{fig:Figure 4}(d) shows one example run for factorizing the 20-bit semiprime $C=1{,}022{,}117=1009\times1013$.

\subsection*{Scaling study on integer factorization}
\label{sec:TTS}

 Finally, we measure how the end-to-end optimization TTS scales with problem size. To separate the framework advantage from the generator advantage, we use proposals generated by long-run MCMC rather than IsingFormer. We vary the multiplier-circuit size and compare how the TTS of PT and TAPT scales as a function of the product bit length (see Methods for further details). As shown in Fig.~\ref{fig:Figure TTS}, exponential fits of the form $\mathrm{TTS}(n)=A\exp(kn)$ yield $k=0.846$ for PT and $k=0.564$ for TAPT. Thus, the fitted TAPT exponent corresponds to an approximately $33\%$ reduction relative to the fitted PT exponent. The scaling improvement therefore does not depend on IsingFormer-generated proposals.

\section*{DISCUSSION}

We asked what happens when PT is augmented with global proposals. Across our experiments, these proposals are generated either by long-run MCMC or by IsingFormer, showing that the framework is not tied to a specific proposal-generation method. This in turn motivates optimizing proposal generation itself. In the present IsingFormer implementation, each spin is represented using a two-token vocabulary; more efficient tokenizations could instead pack multiple spins into a single token, substantially shortening the autoregressive sequence and reducing proposal-generation cost.

Training cost is the other issue, and we address it in two ways. The first is specific to the generative model: IsingFormer learns to interpolate across different $\beta$ values and to perform conditional completion under unseen clamped settings. The second is specific to the problem encoding. For integer factorization, a universal multiplier formulation allows a single trained model to be reused across many problem instances, thereby amortizing the training cost. Although not all problems can be encoded through a universal instance, these results suggest that similar problem formulations may be useful for other families of optimization problems beyond factorization. This amortization does not remove training-data or proposal-bank generation costs, and learned samplers do not automatically overcome hard-sampling barriers~\citep{ciarella2023fails,ghio2024sampling}. These limitations are why we combine learned proposal generators with established MCMC dynamics instead of using either one alone.

\section*{METHODS}

\paragraph{Local Monte Carlo sweeps.}
For an Ising problem with interacting spins $m_i \in \{-1,+1\}$, each spin is updated according to~\cite{aadit2022massively}
\begin{equation}
I_i = \sum_j J_{ij}m_j + h_i
\label{eq:input}
\end{equation}
\begin{equation}
m_i = \operatorname{sgn}\!\left[\tanh(\beta I_i) - r_{[-1,1]}\right]
\label{eq:Gibbs_update}
\end{equation}
where $J_{ij}$ denotes the interaction between spins $i$ and $j$, $h_i$ is the local bias, $\beta$ is the inverse temperature, and $r_{[-1,1]}$ is uniformly sampled from $[-1,1]$. A full MCMC sweep consists of updating all spins once according to Eqs.~\eqref{eq:input} and~\eqref{eq:Gibbs_update}.

\paragraph{Benchmark instances.}
We consider both spin-glass optimization and integer factorization problems. For the spin-glass benchmarks, we use the 3D spin-glass instances introduced and used by D-Wave \citep{king2023quantum}, and for the ground-state energy $E_{\mathrm{gnd}}$, we use the putative ground-state energy reported in~\citep{chowdhury2025pushing}. A total of 300 instances are used for the MCMC benchmarks. For IsingFormer, we use 20 spin-glass instances, with a separate IsingFormer model trained for each instance. For factorization, we map semiprime factorization to Ising models using invertible multiplier circuits. We assume that the two factors, $A$ and $B$, have the same maximum bit width and construct multiplier circuits of different sizes to study semiprimes with different total bit lengths. For a fixed circuit size, $A$ and $B$ may have fewer significant bits than the maximum supported bit width. The product bits are clamped to specify each factorization instance, while the factor bits remain free. For each problem size, we evaluate 440 different semiprime instances.

\paragraph{Graph coloring.}
Instead of updating all spins serially, non-neighboring spins can be updated in parallel. For the 3D spin-glass instances with $L=10$, we partition the supplied interaction graph into independent color classes. For the factorization instances, we perform graph coloring using the DSATUR algorithm~\citep{brelaz1979new}. Five colors are sufficient for all circuit sizes considered in this work. Spins belonging to the same color are non-neighboring and are therefore updated in parallel.

\paragraph{Parallel tempering.}
Our PT implementation uses a temperature schedule designed to achieve approximately uniform swap acceptance between neighboring replicas~\cite{Isakov2015optimising,chowdhury2025pushing}. Once determined, the temperature ladder is kept fixed for all subsequent PT and TAPT simulations. For a given problem family and graph size, the same temperature schedule is used across all instances; for example, all 3D spin-glass instances use the same inverse-temperature ladder. For the actual PT simulations, each local Monte Carlo sweep is followed by replica-exchange attempts between neighboring replicas.

\paragraph{IsingFormer architecture and training.}
We use a decoder-only Transformer to model the distribution of spin configurations. For the ferromagnetic Ising model, we use $d_{\mathrm{model}}=64$, 2 attention heads, 2 Transformer layers, and an FFN dimension of $2d_{\mathrm{model}}$. For the 3D spin-glass and factorization problems, we use a larger model with $d_{\mathrm{model}}=256$, 4 attention heads, 5 Transformer layers, and an FFN dimension of $4d_{\mathrm{model}}$. The models are trained autoregressively on configurations generated by long-run MCMC at multiple temperatures. Further details of the IsingFormer architecture and training parameters are provided in Supplementary Section~\ref{supp: IsingFormersTraining}.

\paragraph{Proposal and dataset generation for TAPT framework.}
Configurations generated by long-run MCMC are used to train IsingFormer at multiple temperatures. Depending on the experiment, global proposals for TAPT are generated either by IsingFormer or directly from long-run MCMC. Except for the contextual ablation in Supplementary Fig.~\ref{fig:replica,context}(b), proposal configurations are generated in advance for each temperature. All PT and TAPT simulations were conducted on an \textit{NVIDIA RTX 6000 Ada} GPU.

\paragraph{Time-to-solution analysis.}
We evaluate optimization performance using time-to-solution (TTS), computed from the measured probability of finding the correct solution after a given number of MCS. For each problem size, we test 440 distinct factorization instances and perform 400 independent runs per instance. The simulation duration is optimized independently for PT and TAPT, and we report the median TTS across the 440 instances at each problem size. To convert TTS from MCS to seconds, we use the measured end-to-end optimization time for each method, conditional on a pre-generated proposal bank. PT timing covers the complete optimization run. TAPT timing additionally includes proposal retrieval and application, proposal-energy evaluation and acceptance or rejection, local Monte Carlo sweeps, replica exchanges, and all other operations performed during optimization, while offline proposal-bank generation is excluded.
\section*{Acknowledgments}

The authors acknowledge support from the Office of Naval Research (ONR) through the ONR-MURI under Grant No. N00014-23-1-2708 and from the National Science Foundation (NSF) through the CAREER Award under Grant No. CCF-2106260. This material is based upon work supported by, or in part by, the Army Research Laboratory under Grant No. W911NF-24-1-0228.

\section*{Author contributions}

 S.B. and K.Y.C. conceived the project. S.B. implemented the TAPT framework, including training the Transformer models, generating proposals, and conducting all optimization experiments and benchmarks. C.D. guided the formulation and initial integration of the TAPT framework for optimization. S.C. performed the free-energy derivation and related simulations. S.B. and A.S.A. developed the transition-matrix formulation and performed the associated analysis. K.L. contributed to the literature review and to relating TAPT to prior work. All authors contributed to discussing the results and writing the manuscript.
\section*{Competing interests}

The authors declare no competing interests.

\section*{Data and code availability}

Data and code necessary to reproduce the main plots will be made available in a public repository upon publication.
Additional data are available upon reasonable request.

\balance
\bibliographystyle{unsrtnat}

\clearpage

\newcommand{\beginsupplement}{%
        \setcounter{subsection}{0}%
        \setcounter{subsubsection}{0}%
        \renewcommand{\thesubsection}{S\arabic{subsection}}%
        \renewcommand{\thesubsubsection}{\thesubsection.\arabic{subsubsection}}%
        \setcounter{table}{0}%
        \renewcommand{\thetable}{S\arabic{table}}%
        \renewcommand{\theHtable}{S\arabic{table}}%
        \setcounter{figure}{0}%
        \renewcommand{\thefigure}{S\arabic{figure}}%
        \renewcommand{\theHfigure}{S\arabic{figure}}%
        \setcounter{equation}{0}%
        \renewcommand{\theequation}{S.\arabic{equation}}%
        \renewcommand{\theHequation}{S.\arabic{equation}}%
     }

\newcolumntype{Z}[1]{S[table-format=#1]}
\onecolumngrid
\begin{center}
    {\Large\bfseries Supplementary Information}
\end{center}
\beginsupplement

\setcounter{subsection}{0}

\subsection{CONTEXT EXPERIMENTS ON 2D ISING}
\label{app:sec_context_exp}

In order to study constrained sampling and boundary-conditioned inference in the 2D Ising model, an exact baseline for the free energy under clamped boundary conditions is required. While the unconstrained Ising free energy has well-known closed-form solutions, the case with a full boundary row (or other partial boundary clamping) in an arbitrary fixed configuration is less standard. Deriving an exact expression for this conditional free energy serves two purposes: (1) it provides a ground-truth benchmark against which approximate MCMC or neural samplers can be evaluated and (2) it highlights the structural modification needed when external clamping is present, namely, the conversion of boundary interactions into effective fields, which can then be absorbed using the ghost-spin trick and expressed via the Kac--Ward/Kasteleyn--Pfaffian theorem \cite{Kac1952,Cimasoni_2010,KASTELEYN19611209,Kardar_2007}.

\subsubsection{EXACT FREE ENERGY EXPRESSION FOR THE CLAMPED EXPERIMENT ON 2D ISING}

In statistical physics, free energy $F$ is defined as
\begin{equation}
F(\beta) = -\frac{1}{\beta} \ln{Z(\beta)}
\end{equation}
where $Z(\beta)$ is the partition function at a given inverse temperature, $\beta$:
\begin{equation}
Z(\beta) = \sum_{\{m\}}\exp{(-\beta E(\{m\}))}
\end{equation}
with $E(\{m\})$ being the energy of the spin configuration $\{m\}$. For the 2D Ising case: 
\begin{equation}
E(\{m\})
=-J\sum_{i=1}^{L_y-1}\sum_{j=1}^{L_x}m_{i,j}m_{i+1,j}
-J\sum_{i=1}^{L_y}\sum_{j=1}^{L_x-1}m_{i,j}m_{i,j+1}
\label{app:context_En}
\end{equation}
Here we consider an $L_y\times L_x$ 2D Ising system with open boundary conditions in all directions, except that the spins on the top boundary (corresponding to $i=L_y$) are clamped to a fixed configuration. There are three categories of spin--spin connections: (1) clamped--clamped connections (the horizontal connections on the top row; we denote the set of clamped spins by $F$), (2) clamped--free connections (the vertical connections between the top row and the row immediately below it), and (3) free--free connections (all other horizontal and vertical connections; we denote the set of free spins by $U$). Let $\sigma_x\in\{\pm1\}$ denote the fixed top spins with a prescribed clamping pattern $\sigma=(\sigma_0,\ldots,\sigma_{L_x-1})$ (e.g. $+,-,-,+$). For simplicity, we rewrite the energy for a spin configuration $m=\{m_i\}_{i\in U}\in\{\pm1\}^U$ as follows:
\begin{equation}
E(m;\sigma)
= -J\!\!\sum_{\langle i,j\rangle\in E(U,U)}\! m_i m_j
  -J\!\!\sum_{\langle x,y\rangle\in E(F,F)}\! \sigma_x \sigma_y
  -J\!\!\sum_{\langle x,j\rangle\in E(F,U)}\! \sigma_x m_j ,
\end{equation}
where $E(A,B)$ collects edges with one endpoint in $A$ and the other in $B$.
The conditional partition function that \emph{sums only over the free spins} is
\begin{equation}
Z(\beta;\sigma)
=\sum_{m\in\{\pm1\}^U} \exp{\!\bigl(-\beta\,E(m;\sigma)\bigr)}
\end{equation}
We are interested in the free energy of this system with the clamped row on top, i.e.,  $F(\beta;\sigma)=-\beta^{-1}\log Z(\beta;\sigma)$. Let us now define
\begin{equation}
Z_1 = \exp{(\beta J \!\!\sum_{\langle x,y\rangle\in E(F,F)}\! \sigma_x \sigma_y)}
\end{equation}
This contains the contribution of the clamped row in $Z$ and is a constant with respect to free spin configurations, $m$. This yields
\begin{align}
\label{eq:after-factor}
Z(\beta;\sigma) &= Z_1\,
\prod_{e\in E(U,U)}\!\cosh{(\beta J)}\,
\prod_{\langle x,j\rangle\in E(F,U)}\!\cosh{(\beta J)}\,\nonumber\\
&\sum_{m\in\{\pm1\}^U} \;\prod_{\langle i,j\rangle\in E(U,U)} \!\Bigl(1+m_i m_j \tanh{(\beta J)}\Bigr)
\prod_{\langle x,j\rangle\in E(F,U)} \!\Bigl(1+\sigma_x m_j \tanh{(\beta J)}\Bigr)
\end{align}
where we have made use of the identity 
\begin{equation}
{\rm e}^{Km_im_j}
=
\cosh(K)\left(1+m_im_j\tanh(K)\right)
\end{equation}
for bipolar variables $m_i$ and $m_j$. Next we apply an important trick, we re-write the partition function as
\begin{align}
\label{eq:after-factor2}
Z(\beta;\sigma) &= Z_1\,
\prod_{e\in E(U,U)}\!\cosh{(\beta J)}\,
\prod_{\langle x,j\rangle\in E(F,U)}\!\cosh{(\beta J)}\,\nonumber\\
&\frac{1}{2}\sum_{g\in\{\pm1\}}\sum_{m \in\{\pm1\}^U} \;\prod_{\langle i,j\rangle\in E(U,U)} \!\Bigl(1+m_i m_j \tanh{(\beta J)}\Bigr)
\prod_{\langle x,j\rangle\in E(F,U)} \!\Bigl(1 + g\,m_j \tanh{(\beta J \sigma_x)}\Bigr)
\end{align}
where we have replaced the clamped--free factors with free--ghost factors by introducing a single `ghost' spin $g\in\{+1,-1\}$. This transformation is exact: the $g=+1$ sector reproduces the original fixed-boundary problem, while the $g=-1$ sector is equivalent under a global flip of the free spins. The prefactor $1/2$ therefore removes this duplication. The double sum in Eq.~(\ref{eq:after-factor2}) can be written as
\begin{align}
&\sum_{g=\pm1}\sum_{m\in\{\pm1\}^U}
\prod_{\langle i,j\rangle\in E(U,U)} \!\Bigl(1+m_i m_j \tanh{(\beta J)}\Bigr)
\prod_{\langle x,j\rangle\in  E(F,U)} \!\Bigl(1+g\,m_j\, \tanh{(\beta J \sigma_x)}\Bigr) \nonumber\\
&=\; 2^{\,|U|+1}\;\Xi
\end{align}
For the boundary-clamping geometry used here, the ghost can be placed in the exterior face and connected to the interface vertices without crossings, so the augmented graph remains planar. The Kac--Ward theorem~\cite{Kac1952,Cimasoni_2010} then gives
\begin{equation}
\Xi \;=\; \sqrt{\det{\!\bigl(I - Q\bigr)}} ,
\end{equation}
with $Q$ being the Kac-Ward transfer matrix indexed by directed edges $e=(u\!\to\!v)$ of the Ising lattice with the ghost spin $g$. Assuming two edges, $e=(u\!\to\!v)$ and $e'=(v\!\to\!w)$, the elements of the $Q$ matrix are written as
\begin{equation}
\label{eq:T-def}
Q_{e,e'} \;=\;
\begin{cases}
C_K\,\exp{\!\bigl(\frac{i}{2}\,\Delta\theta(e,e')\bigr)},
&\text{if } u \neq w,\\[4pt]
0, &\text{otherwise.}
\end{cases}
\end{equation}
Here $\Delta\theta(e,e')$ is the turning angle from edge $e$ to $e'$
under a fixed planar embedding, $C_K = \tanh{(\beta J)}$ for connections between free spins and $C_K = \tanh{(\beta J \sigma_x)}$ for free spin-ghost connections.

Combining the pieces in Eq.~(\ref{eq:after-factor2}) we obtain the exact polynomial-time
expression for free energy:
\begin{align}
\label{eq:Z-final}
-\beta F &=
\log Z(\beta;\sigma) \nonumber \\
&= \sum_{\langle x,y\rangle\in E(F,F)} \beta J\,\sigma_x\sigma_y+ \sum_{e\in E(U,U)} \log\cosh{(\beta J)}
+\sum_{\langle x,j\rangle\in E(F,U)} \log\cosh{(\beta J)} \nonumber\\
&+ |U|\,\log 2
+ \frac12\,\log\det{\!\bigl(I - Q\bigr)}
\end{align}

Finally, it can be noted that in the case of our experiment on 2D Ising with semicircular defect, all rows above the last clamped row of the top half also contribute to a constant term in the energy and hence can be included as a constant product term with the partition function (or as a sum in $\log{Z}$) expression obtained above. 

\textbf{Numerical estimation of free energy.} Direct estimation of $Z$ from samples is hard because of the huge size of the state space. Hence, to estimate free energy at a given inverse temperature, we do the following:

\begin{align}
\frac{\partial (\beta F)}{\partial \beta} &=  -\frac{\partial \ln{(Z)}}{\partial \beta}=-\frac{1}{Z}\,\frac{\partial Z}{\partial \beta}\\
-\frac{1}{Z}\,\cfrac{\partial Z}{\partial \beta} &= \sum_{\{m\}} E(\{m\})\frac{\exp{(-\beta E(\{m\}))}}{Z} = E_{\rm{avg}}(\beta)
\end{align}

We first estimate $E_{\rm{avg}}(\beta)$ from samples and then integrate over $\beta$:
\begin{align}
\beta F(\beta) = {\rm const.} + \int_{0}^{\beta} E_{\rm{avg}}(\omega)  \,{\rm d}\omega
\end{align}
The integration constant is found by evaluating $-\ln Z$ at $\beta=0$. Because the conditional partition function sums only over the free spins, $Z(0;\sigma)=2^{|U|}$ and the integration constant is $-|U|\ln 2$.

In this work, we do this numerical integration by linearly dividing the range from $10^{-3}$ to $\beta$ into 25 segments, taking 100 samples at each of the intermediate inverse temperatures and using trapezoidal rule for integration.

The second derivative of $\beta F$ also contains important information about the system and is related to the variance of energy:
\begin{align}
\frac{\partial^2 (\beta F)}{\partial \beta^2} &=  +\,\left(\frac{1}{Z}\frac{\partial Z}{\partial \beta}\right)^2-\frac{1}{Z}\,\frac{\partial^2 Z}{\partial \beta^2}\nonumber\\
&=  +(E_{\rm avg}(\beta))^2-\sum_{\{m\}} E^2(\{m\})\frac{\exp{(-\beta E(\{m\}))}}{Z}\nonumber\\
&= -\text{var}(E(\beta))
\end{align}

\begin{figure}[t!]
    \centering
    \includegraphics[width=0.99\textwidth, keepaspectratio]{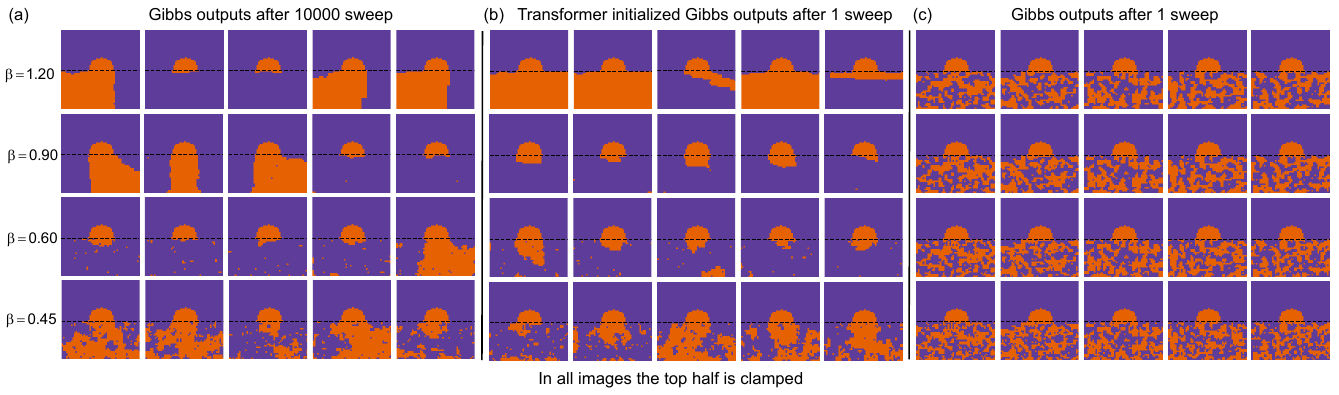}
    \caption{\textbf{Effect of transformer initialization under half-clamped boundary conditions:} Representative spin configurations on a 50 $\times$ 50 Ising grid with the top half clamped with a semicircular defect (orange: \(-1\), purple: \(+1\)) at different inverse temperatures \(\beta\). (a) Configurations after \(10^4\) Gibbs sweeps, used here as the long-run reference. (b) Transformer-initialized Gibbs sampling after one sweep produces configurations that qualitatively resemble the long-run reference. (c) By contrast, Gibbs sampling from random initialization after one sweep does not reproduce the same boundary-induced structure. The visual resemblance in (b) demonstrates improved initialization, but does not by itself constitute a convergence diagnostic.}
    \label{fig:collage_clamped}
\end{figure}

In Table~\ref{tab:free_energy_clamped_obc_L50}, we show the comparison of free energy estimates from Gibbs sampler and transformer with exact free energy value at several $\beta$. Supplementary Fig.~\ref{fig:collage_clamped} shows sample configurations for the clamped 2D Ising example discussed in the main text.

\begin{table}[!t]
\centering
\begin{threeparttable}
\caption{Free energy per spin $F/L^2$ (top half clamped, open  boundary at all sides, $L=50$) at several inverse temperatures $\beta$. Estimates from transformers are averages of 100 samples. Gibbs estimates are computed from the averages of 100 independent runs.}
\label{tab:free_energy_clamped_obc_L50}
\setlength{\tabcolsep}{5pt}
\renewcommand{\arraystretch}{1.3}
\begin{tabular}{
  Z{1.2}   
  Z{-1.2}  
  Z{-1.2}  
  Z{-1.2}  
  Z{-1.2}  
  Z{-1.2}  
  Z{-1.2}  
}
\toprule
& \multicolumn{6}{c}{\textbf{Free energy per spin}, $F/L^2$} \\
\cmidrule(lr){2-7}
{\bfseries $\beta$} &
\multicolumn{1}{c}{\makecell{Exact}} 
&
\multicolumn{1}{c}
{\makecell{Transformer}} 
&
\multicolumn{1}{c}{\makecell{Gibbs\\(10 sweeps)}} 
&
\multicolumn{1}{c}{\makecell{Gibbs\\($10^2$ sweeps)}} 
&
\multicolumn{1}{c}{\makecell{Gibbs\\($10^3$ sweeps)}} 
&
\multicolumn{1}{c}{\makecell{Gibbs\\($10^4$ sweeps)}}\\
\midrule
1.2 & -1.919794 & -1.916203 & -1.826265 & -1.909454 & -1.911647 & -1.911051 \\
0.8 & -1.921940 & -1.925591 & -1.852581 & -1.914358 & -1.917699 & -1.916191 \\
0.2 & -2.74415  & -2.765629 & -2.738982 & -2.738322 & -2.740267 & -2.738134 \\
\bottomrule
\end{tabular}

\end{threeparttable}
\end{table}

\subsection{TAPT: DETAILED BALANCE AND TRANSITION-MATRIX ANALYSIS}
\label{TAPT derivation}
\begin{figure}[t!]
    \centering
    \includegraphics[width=\linewidth]{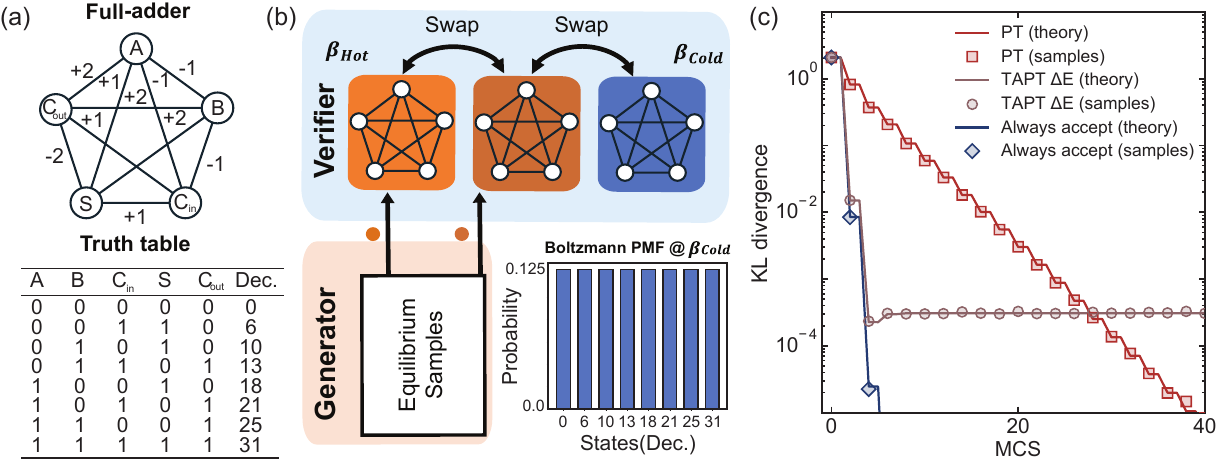}
    \caption{\textbf{Proposal acceptance based on the \(\Delta E\) rule in a full-adder Ising system.} \textbf{(a)} Five-spin Ising encoding of a full-adder circuit and its corresponding truth table. \textbf{(b)} Controlled three-replica PT system in which independent exact-equilibrium proposals are supplied to the first two replicas at their corresponding inverse temperatures. No proposals are supplied to the coldest replica. The inset shows the target Boltzmann distribution of the coldest replica. \textbf{(c)} KL divergence \(D_{\mathrm{KL}}\!\left(p_{\mathrm{cold}} \,\|\, \pi_{\beta_{\mathrm{cold}}}\right)\) between the cold-replica distribution and its target Boltzmann distribution as a function of MCS. Theory denotes the exact transition-matrix calculation, while symbols show Monte Carlo estimates. Standard PT and the always-accept proposal rule converge toward the target distribution. In contrast, applying the \(\Delta E\)-only acceptance rule to exact equilibrium proposals results in a nonzero asymptotic KL divergence. The three replicas use inverse temperatures $\beta=\{0.5,1.7,4.6\}$.}
    \label{fig:TAPT_understanding}
\end{figure}

In this section, we discuss the mathematical details of the TAPT framework and validate our claims through numerical and transition-matrix analysis on a small five-spin full-adder example, as shown in Fig.~\ref{fig:TAPT_understanding}. We show that using only the $\Delta E$ Metropolis criterion for $P_{\text{accept}}$ does not preserve detailed balance for the proposal moves considered here and is therefore not suitable for unbiased equilibrium sampling. Nevertheless, it remains effective for ground-state search in the optimization experiments considered in this work.\newline

\noindent \textbf{Part 1: Proposal moves and detailed balance.}

\noindent Consider a Markov chain over spin configurations \(x\), with target Boltzmann
distribution
\begin{equation}
    \pi_\beta(x) = \frac{1}{Z_\beta}\exp[-\beta E(x)] .
\end{equation}
For a proposal-based MCMC move, the transition from \(x\) to \(x'\) is split
into a proposal probability and an acceptance probability:
\begin{equation}
    P(x'|x) = q_\beta(x'|x) A_\beta(x\to x'), \qquad x'\neq x.
\end{equation}
Detailed balance with respect to \(\pi_\beta\) requires, for every pair
\(x,x'\),
\begin{equation}
    \pi_\beta(x) q_\beta(x'|x) A_\beta(x\to x')
    =
    \pi_\beta(x') q_\beta(x|x') A_\beta(x'\to x).
    \label{eq:db}
\end{equation}
Therefore, the acceptance probabilities must satisfy
\begin{equation}
    \frac{A_\beta(x\to x')}{A_\beta(x'\to x)}
    =
    \frac{\pi_\beta(x') q_\beta(x|x')}
         {\pi_\beta(x) q_\beta(x'|x)} 
    \label{eq:ratio}
\end{equation}
The standard Metropolis--Hastings choice is
\begin{equation}
    A_\beta(x\to x')
    =
    \min\left[
        1,\,
        \frac{\pi_\beta(x')}
             {\pi_\beta(x)}
        \cdot
        \frac{q_\beta(x|x')}{q_\beta(x'|x)}
    \right]
    \label{eq:mh}
\end{equation}
The second factor is the Hastings correction. It is unnecessary only when
the proposal is symmetric,
\(q_\beta(x'|x)=q_\beta(x|x')\). Learned proposals need not be symmetric,
so any proposal asymmetry must be accounted for in the acceptance rule.

In the ideal setting, the proposal mechanism gives an exact equilibrium
sample at the same inverse temperature \(\beta\). That is, the proposed
state is independent of the current state and is distributed according
to the target:
\begin{equation}
    q_\beta(x'|x)
    =
    q_\beta(x')
    =
    \pi_\beta(x'),
    \qquad
    q_\beta(x|x')
    =
    q_\beta(x)
    =
    \pi_\beta(x).
    \label{eq:idealq}
\end{equation}
Substituting these expressions into the Metropolis--Hastings rule
in Eq.~\eqref{eq:mh} gives
\begin{align}
    A_\beta(x\to x')
    &=
    \min[1,1]
    =
    1 .
\end{align}
Thus, an exact equilibrium proposal should always be accepted
(blue curve in Fig.~\ref{fig:TAPT_understanding}(c)). Accordingly,
our choice of $P_{\text{accept}}$ does not preserve detailed balance
even in the ideal setting, as indicated by the brown curve in
Fig.~\ref{fig:TAPT_understanding}(c).

\noindent \textbf{Part 2: Transition matrix.}

\noindent \textbf{First, the swap dynamics}. We construct the transition matrix $W_{\mathrm{swap}}$ that describes the swap transition between two neighboring replicas $(r,r+1)$. Each replica state is an $N$-spin configuration $S_r=(m_1,\ldots,m_N)\in\{-1,+1\}^N$, where $m_i$ denotes the state of spin $i$ in replica $r$.
Therefore, a two-replica state is written as $(S_r,S_{r+1})$, and the corresponding state space has size $2^N\times 2^N=2^{2N}$.

Although
$W_{\mathrm{swap}}\in\mathbb{R}^{2^{2N}\times 2^{2N}}$,
it is highly sparse. There are only $2^{2N}-2^N$ possible
off-diagonal swap transitions. This is because, from a given state
$(S_r,S_{r+1})$, the swap move can only either exchange the two
replica configurations or leave them unchanged. In particular, when
$S_r=S_{r+1}$, swapping does not change the state. The swap acceptance
probability is
\[
P_{\mathrm{swap}}
=
A_r(S_r,S_{r+1})
=
\min\left\{
1,\,
\exp\left[
(\beta_{r+1}-\beta_{r})
\left(E(S_{r+1})-E(S_{r})\right)
\right]
\right\}.
\]

We can write the entries of the $W_{\mathrm{swap}}$ matrix as
\[
W_{\mathrm{swap}}
\big[
(S'_r,S'_{r+1})\leftarrow(S_r,S_{r+1})
\big]
=
\begin{cases}
1,
& S_r=S_{r+1},
\quad
(S'_r,S'_{r+1})=(S_r,S_{r+1}),
\\[4pt]
A_r(S_r,S_{r+1}),
& S_r\neq S_{r+1},
\quad
(S'_r,S'_{r+1})=(S_{r+1},S_r),
\\[4pt]
1-A_r(S_r,S_{r+1}),
& S_r\neq S_{r+1},
\quad
(S'_r,S'_{r+1})=(S_r,S_{r+1}),
\\[4pt]
0,
& \text{otherwise}.
\end{cases}
\]
This matrix is column stochastic. For each input state, the probability
either remains entirely in the same state when $S_r=S_{r+1}$, or is
split between accepting and rejecting the swap when
$S_r\neq S_{r+1}$. \newline

\noindent \textbf{Second, the proposal of equilibrium samples}.
Similar to the discussion of Eq.~\eqref{eq:idealq}, an exact equilibrium
proposal at inverse temperature $\beta$ is independent of the current
state and is distributed according to the Boltzmann distribution:
\[
q_\beta(x'|x)=\pi_\beta(x').
\]
Therefore, when all proposals are accepted, the proposal transition
matrix can be written as
\begin{equation}
    W_{\mathrm{prop},\beta}
    =
    \boldsymbol{\pi}_\beta \mathbf{1}^{T},
    \label{eq:prob}
\end{equation}
where $\boldsymbol{\pi}_\beta=
[\pi_\beta(x_1),\ldots,\pi_\beta(x_{2^N})]^T$ is the Boltzmann probability vector and $\mathbf{1}$ is the $2^N$-dimensional all-ones vector. Thus, every column of $W_{\mathrm{prop},\beta}$ is equal to the Boltzmann distribution. For the example in Fig.~\ref{fig:TAPT_understanding}(b), proposals are applied only to the first two replicas, while no proposals are supplied to the coldest replica at $\beta_{\mathrm{cold}}$.

When proposals are accepted according to the $\Delta E$ Metropolis rule, the off-diagonal transition probabilities become
\[
W_{\Delta E,\beta}(x'\leftarrow x)
=
\pi_\beta(x')A_{\Delta E}(x\to x'),
\qquad x'\neq x,
\]
where
\[
A_{\Delta E}(x\to x')
=
\min\left[
1,\,
\exp\left(-\beta[E(x')-E(x)]\right)
\right]
\]
The rejected probability is assigned to the diagonal element,
\[
W_{\Delta E,\beta}(x\leftarrow x)
=
1-
\sum_{x'\neq x}
\pi_\beta(x')A_{\Delta E}(x\to x')
\]
so that each column of the transition matrix sums to one.

\noindent \textbf{Third, the sweep dynamics}.
For a single-site update of spin $m_i$, we use the Gibbs rule
\[
P(m_i=+1\mid I_{i})
=
\frac{1+\tanh(\beta I_i)}{2},
\qquad
m_i\in\{-1,+1\},
\]
where the local field is
\[
I_i
=
\left(
\sum_{j\neq i}J_{ij}m_j+h_i
\right)
\]

Following the transition-matrix construction discussed
in~\cite{abdelrahman2025generalized}, we define $W_i$ as the transition
matrix associated with a Gibbs update of spin $i$. A complete sequential
sweep over all $N$ spins is then described by
\begin{equation}
    W_{\mathrm{sweep}}
    =
    W_N W_{N-1}\cdots W_1 
\end{equation}
At finite temperature, each Gibbs update assigns nonzero probability
to both spin values, allowing the sweep dynamics to connect the
configuration space. Similarly, $W_{\mathrm{prop},\beta}$ has strictly
positive entries because $\pi_\beta(x)>0$ for every state. These
properties ensure ergodicity of the combined dynamics considered here.
\newline

The three transition matrices discussed above are used to construct the
theoretical results shown in Fig.~\ref{fig:TAPT_understanding}. For PT,
we use $W_{\mathrm{sweep}}$ and $W_{\mathrm{swap}}$. For TAPT with
always-accepted equilibrium proposals (labeled as ``always accept''),
we use the proposal transition matrix $W_{\mathrm{prop},\beta}$ defined
in Eq.~\eqref{eq:prob}. When proposals are instead accepted according
to the $\Delta E$ Metropolis rule, we use the corresponding transition
matrix $W_{\Delta E,\beta}$ defined above. As shown in
Fig.~\ref{fig:TAPT_understanding}, the $\Delta E$-only acceptance rule
does not preserve detailed balance when applied to exact equilibrium
proposals.

\begin{figure*}[t!]
    \centering
    \includegraphics[width=0.98\linewidth, keepaspectratio]{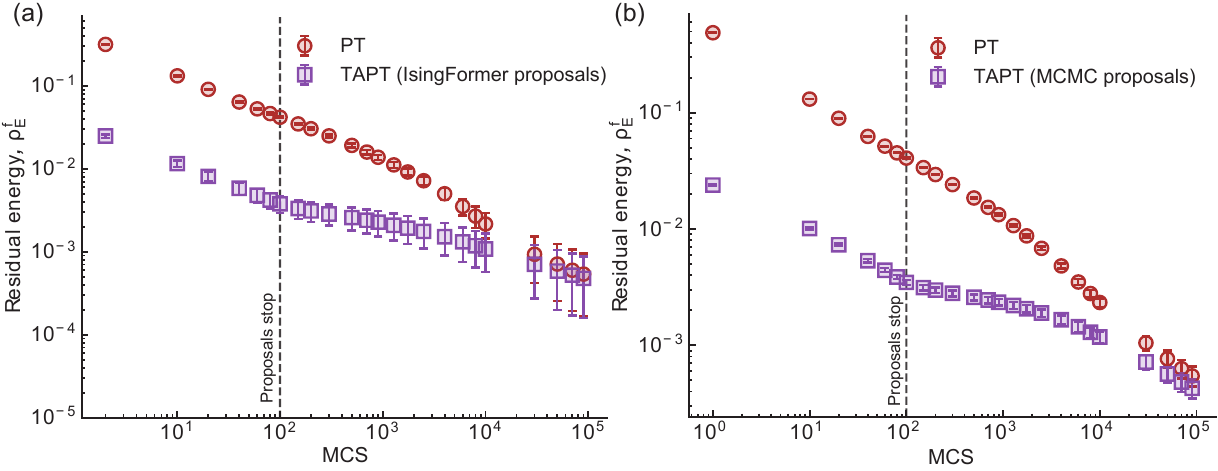}
    \caption{\textbf{The TAPT advantage diminishes after proposal moves are disabled.} Residual energy is shown as a function of MCS for 3D spin-glass instances (\(L=10\)). TAPT proposal moves are applied only during the first 100 MCS, with the cutoff indicated by the vertical dashed line. After proposals are disabled, the TAPT curves progressively approach the PT baseline, indicating that the advantage gained during the proposal phase is not sustained without continued proposal moves. \textbf{(a)} IsingFormer-generated proposals, evaluated on 20 spin-glass instances. \textbf{(b)} Proposals drawn from long-run MCMC configurations, evaluated on 300 spin-glass instances. Each instance is evaluated using 400 independent runs. Markers show the mean across instances, and error bars indicate 95\% bootstrap confidence intervals.}
    \label{fig:TAPT_discontinuous_proposal}
\end{figure*}

\begin{figure*}[b!]
    \centering
    \includegraphics[width=0.98\linewidth, keepaspectratio]{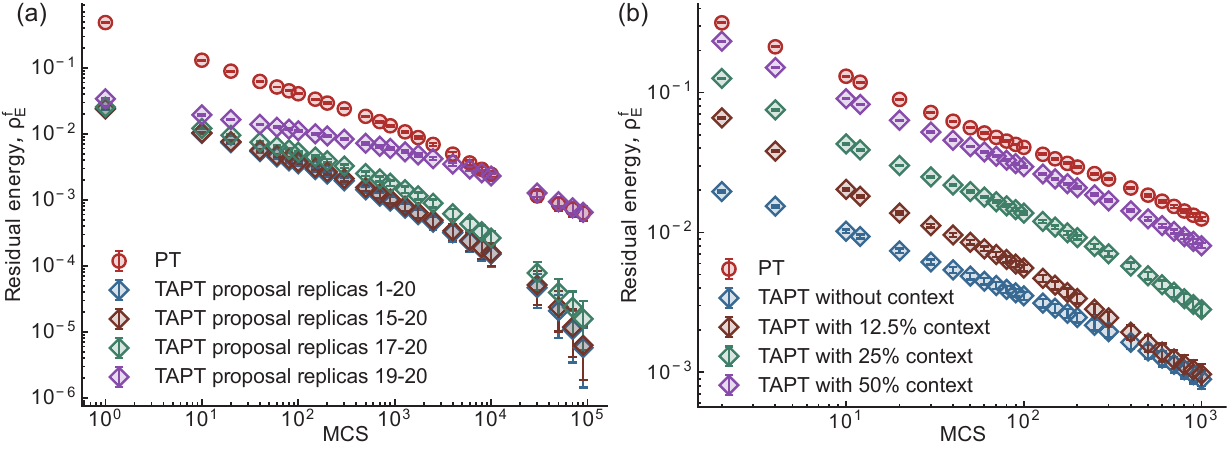}
       \caption{\textbf{Ablation study of proposal coverage and context in TAPT.} Residual energy is shown as a function of MCS for 300 3D spin-glass instances; proposals are generated using long-run MCMC, and no IsingFormer-generated proposals are used. \textbf{(a)} Proposal-coverage ablation. The set of replicas receiving proposals is varied among replicas 1--20, 15--20, 17--20, and 19--20 and compared with baseline PT. Replica indices increase from the hottest toward the coldest end of the 22-replica ladder. \textbf{(b)} Context ablation. Contextual proposals are generated online by randomly selecting 12.5$\%$, 25$\%$, or 50$\%$ of the spins from the current replica, clamping them to their current values, and generating the remaining spins using MCMC under these clamped conditions. The no-context case uses unconstrained MCMC proposals. These experiments isolate the effects of proposal coverage across the temperature ladder and of conditioning proposals on the current replica state. Each instance is evaluated over 400 independent runs, and error bars indicate 95$\%$ bootstrap confidence intervals.}
    \label{fig:replica,context}
\end{figure*}

\subsection{ABLATION AND CONTEXT STUDY}
\label{Supplementary:context}

In contrast to the main 3D spin glass experiment (Fig. \ref{fig: 3d spin glass} in the main text), we investigate the effect of using proposals only as a warm start for the PT ladder. We apply proposals after every MCS during the first 100 MCS, after which TAPT transitions to normal PT dynamics, effectively removing the proposal loop from Algorithm~\ref{alg:augPT}. Our results in Fig.~\ref{fig:TAPT_discontinuous_proposal} show that the proposals initially produce a sharp drop in the \emph{residual energy}, $\rho_{E} = \langle E - E_{\rm gnd} \rangle/N$, compared with PT. However, after terminating the proposals, relying only on standard PT dynamics does not appear to preserve the initial improvement in \emph{residual energy} provided by TAPT, and the residual energy gradually approaches the standard PT baseline without a warm start. Similar to the experiment in Fig. \ref{fig: 3d spin glass} of the main text, we test the same setting using proposals generated by both IsingFormer and long-run MCMC, and both exhibit a similar response after the proposals are terminated. These results suggest that, in this setting, continued proposal moves are needed to sustain the \emph{residual energy} advantage observed with TAPT. \newline

\begin{figure}[t]
\includegraphics[width=0.5\linewidth, keepaspectratio]{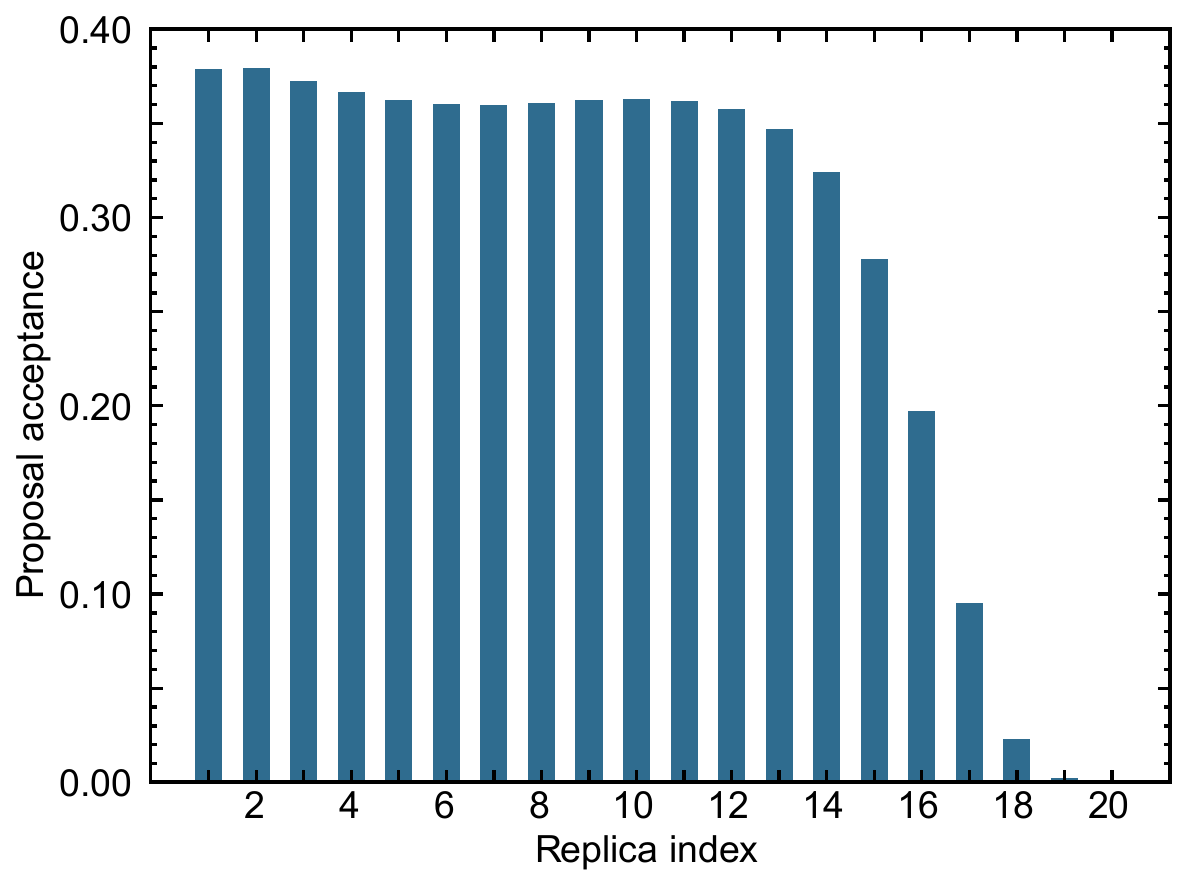}
\caption{\textbf{Proposal acceptance across the PT ladder for 3D spin-glass instances.} The mean proposal acceptance probability is shown for proposal-enabled replicas 1--20 of the 22-replica ladder for TAPT using MCMC-generated proposals, aggregated over the 300 tested instances. Replica indices increase from the hottest toward the coldest end of the ladder. The two coldest replicas, 21 and 22, do not receive proposals.}
\label{fig:proposals acceptance}
\end{figure}

\begin{figure}[t]
\includegraphics[width=0.55\linewidth, keepaspectratio]{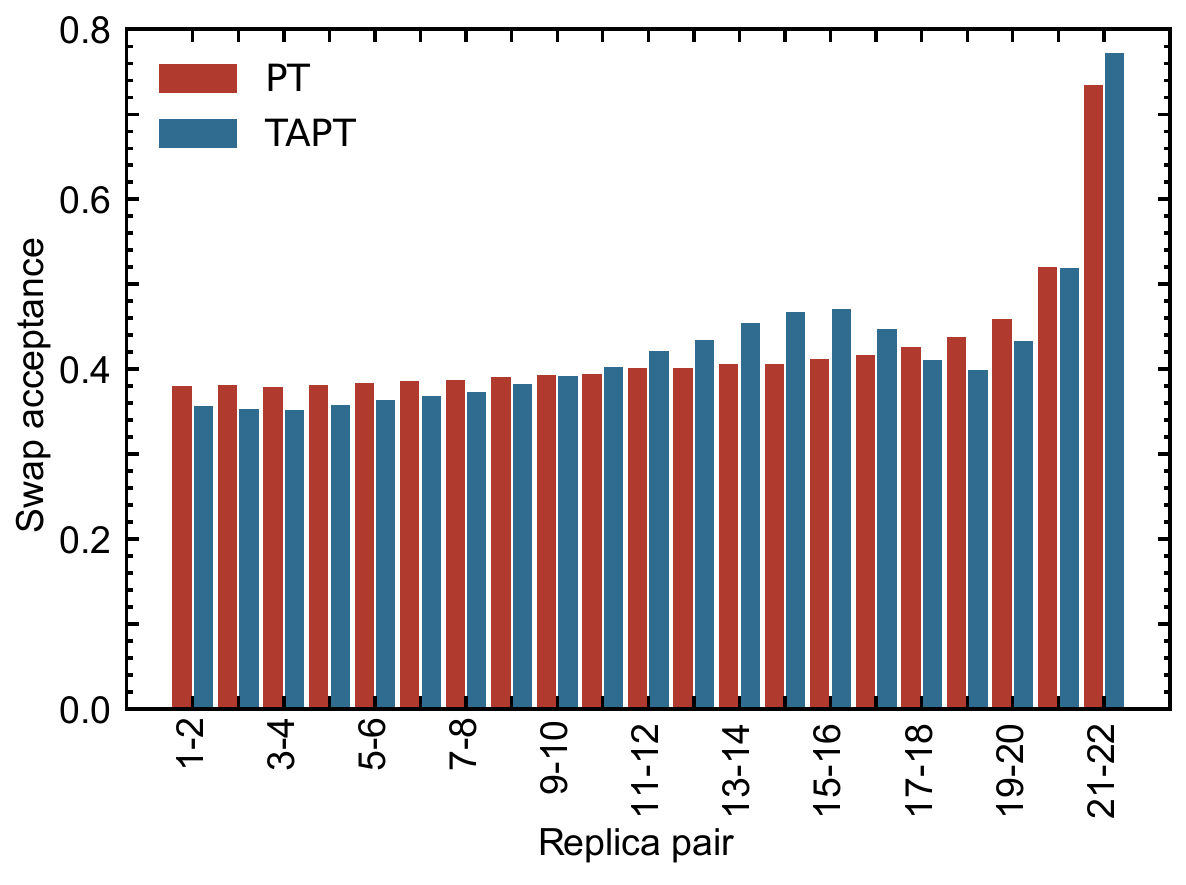}
\caption{\textbf{Swap acceptance across the PT ladder for 3D spin-glass instances.} The mean swap acceptance probability is shown for each adjacent replica pair in the 22-replica ladder for PT and TAPT, using the same 300-instance ensemble as in Figure~S5. Results combine the alternating even--odd and odd--even swap stages. The PT and TAPT swap-acceptance profiles are broadly similar across the PT ladder.}
\label{fig:swap acceptance}
\end{figure}

We then separately investigate the effect of varying the temperature range of the proposals and the effect of providing partial-state context to the proposal generator. Specifically, we vary the proposal-enabled set $\mathcal{R}_{\mathrm{prop}}$ from Algorithm~\ref{alg:augPT}. Fig.~\ref{fig:replica,context}(a) shows the results for replica ranges 19--20, 17--20, 15--20, and 1--20. The results indicate that restricting proposals to only the highest-$\beta$ proposal-enabled replicas provides a smaller \emph{residual energy} advantage, while extending the proposal range to include intermediate-$\beta$ replicas substantially improves performance. Extending the proposals to all proposal-enabled replicas (1--20) does not provide further improvement. The limited benefit at the cold end of the proposal-enabled range is consistent with the low proposal acceptance probabilities observed for these replicas in Fig.~\ref{fig:proposals acceptance}. Notably, we observe broadly similar replica swap-acceptance profiles between PT and TAPT, as shown in Fig.~\ref{fig:swap acceptance}.

We also vary the amount of context provided by the replicas to the generator, corresponding to a fraction of the current replica state. For the contextual cases, proposals are generated online: at each proposal move, a randomly selected subset of spins is clamped to its current replica values, and the remaining spins are generated using MCMC under these clamped conditions. The resulting proposal is then accepted or rejected according to the same optimization-biased $\Delta E$-only rule for $P_{\mathrm{accept}}$. As shown in Fig.~\ref{fig:replica,context}(b), reducing the amount of context consistently improves performance, with context-free proposals providing the largest reduction in \emph{residual energy}. This suggests that current-replica conditioning is not beneficial here.

\begin{figure*}[t]
\includegraphics[width=0.8\linewidth, keepaspectratio]{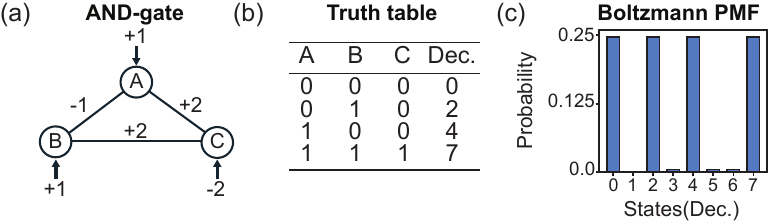}
\caption{
\textbf{Boltzmann distribution of the three-spin AND-gate Ising model.}
\textbf{(a)} Ising representation of the AND gate using three spins,
$A$, $B$, and $C$, with the indicated local fields and pairwise couplings. \textbf{(b)} Corresponding truth table, where the decimal index labels the three-spin state $(A,B,C)$ in binary order. \textbf{(c)} Boltzmann probability distribution over all eight spin states at $\beta=1$. The four logical states satisfying $C=A\land B$, corresponding to decimal states 0, 2, 4, and 7, have the largest Boltzmann probability, while the remaining states are energetically suppressed.
}
\label{fig:AND Gate}
\end{figure*}

\subsection{FACTORIZATION CIRCUIT WITH INVERTIBLE LOGIC GATES}
\label{supp: Invertible_Logic_Gates}

We detail the invertible multiplier used for the integer factorization experiment shown in Fig.~\ref{fig:Figure 4}(a). The circuit is constructed from invertible AND gates (Fig.~\ref{fig:AND Gate}) and full-adder (FA) units (Fig.~\ref{fig:TAPT_understanding}(a)), each represented by coupled Ising spins with coupling coefficients taken from Ref.~\cite{aadit2022massively}. For spins $m_i\in\{-1,+1\}$, logical states are mapped as $0\rightarrow -1$ and $1\rightarrow +1$. We use the Ising energy convention
\begin{equation}
E(\mathbf{m})
=
-\sum_{i<j} J_{ij}m_i m_j
-\sum_i h_i m_i 
\end{equation}
where $J$ is a symmetric interaction matrix with zero diagonal and $h$ contains the local fields.

For the full adder, using the spin ordering
$\mathbf{m}_{\mathrm{FA}}=(A,B,C_{\mathrm{in}},S,C_{\mathrm{out}})$,
the interaction matrix and local fields are
\begin{equation}
J_{FA} =
\begin{pmatrix}
0  & -1 & -1 & +1 & +2 \\
-1 & 0  & -1 & +1 & +2 \\
-1 & -1 & 0  & +1 & +2 \\
+1 & +1 & +1 & 0  & -2 \\
+2 & +2 & +2 & -2 & 0
\end{pmatrix},
\qquad
h_{FA} =
\begin{pmatrix}
0 & 0 & 0 & 0 & 0
\end{pmatrix}.
\end{equation}

For the AND gate, using
$\mathbf{m}_{\mathrm{AND}}=(A,B,C)$,
the corresponding parameters are
\begin{equation}
J_{\mathrm{AND}} =
\begin{pmatrix}
0  & -1 & +2 \\
-1 & 0  & +2 \\
+2 & +2 & 0
\end{pmatrix},
\qquad
h_{\mathrm{AND}} =
\begin{pmatrix}
+1 & +1 & -2
\end{pmatrix}.
\end{equation}

In forward mode, the circuit performs multiplication, $C=A\times B$, by clamping the corresponding input bits $A$ and $B$. During factorization, we instead operate the circuit in reverse: the product $C$ is clamped while $A$ and $B$ remain free, such that ground states of the resulting Ising model encode valid factor pairs. For IsingFormer training, however, the multiplier is operated in its universal free-running configuration, with no product instance imposed. Configurations generated by long-run MCMC from this universal circuit are used to train a single IsingFormer, which is subsequently conditioned on the desired product $C$ during inference.

\subsection{TIME-TO-SOLUTION SCALING FOR INTEGER FACTORIZATION}
\label{supp:time-to-solution}

We evaluate the scaling of PT and TAPT on integer factorization using the time-to-solution (TTS) methodology of \cite{ronnow2014defining}. We report end-to-end optimization TTS conditional on pre-generated proposal banks. For this scaling study, TAPT uses proposals drawn from long-run MCMC configurations rather than IsingFormer-generated proposals, allowing us to isolate the scaling of the TAPT framework from the performance of a particular learned generator.

The scaling study considers multiplier circuits with maximum product sizes of 16, 18, 20, 22, and 24 bits. For each problem size, 440 semiprime instances are evaluated, with the empirical success probability for each instance estimated from 400 independent runs. In TAPT, proposal moves are attempted once per MCS using pre-generated long-run MCMC proposals at the corresponding replica temperatures. For each circuit size, the inverse-temperature ladder is obtained using the same adaptive procedure described in the main text and is subsequently fixed for the PT and TAPT comparisons.

For a single independent run with success probability $s$, the probability of failing to find a solution after $R$ independent repetitions is
\begin{equation}
    P_{\mathrm{fail}}=(1-s)^R
\end{equation}
such that the probability of obtaining at least one successful solution is
\begin{equation}
    P_{\mathrm{target}}
    =
    1-(1-s)^R 
    \label{eq:supp-tts-success}
\end{equation}
The number of independent repetitions required to reach a target success probability $P_{\mathrm{target}}$ is therefore
\begin{equation}
    R(P_{\mathrm{target}})
    =
    \left\lceil
    \frac{\log(1-P_{\mathrm{target}})}
         {\log(1-s)}
    \right\rceil 
    \label{eq:supp-tts-repetitions}
\end{equation}
Throughout this work, we use $P_{\mathrm{target}}=0.99$.

For a run length of $t_a$ MCS, with empirical success probability $s(t_a)$, the corresponding TTS in units of MCS is
\begin{equation}
    \mathrm{TTS}_{\mathrm{MCS}}(t_a)
    =
    t_a
    \left\lceil
    \frac{\log(1-0.99)}
         {\log[1-s(t_a)]}
    \right\rceil 
    \label{eq:supp-tts-mcs}
\end{equation}
Following \cite{ronnow2014defining}, we evaluate the TTS over a range of run lengths and independently optimize $t_a$ for PT and TAPT. The reported TTS is therefore
\begin{equation}
    \mathrm{TTS}_{\mathrm{MCS}}^{\star}
    =
    \min_{t_a}
    \mathrm{TTS}_{\mathrm{MCS}}(t_a)
    \label{eq:supp-tts-optimized}
\end{equation}
The optimal run length is selected independently for each method and circuit size.

After optimizing the run length in MCS, we convert the corresponding computational effort to seconds. Let $\overline{\Delta t}_{\mathrm{MCS}}$ denote the measured average optimization time per trajectory per MCS for the selected run length. The optimized TTS in seconds is then
\begin{equation}
    \mathrm{TTS}_{\mathrm{sec}}^{\star}
    =
    \mathrm{TTS}_{\mathrm{MCS}}^{\star}
    \times
    \overline{\Delta t}_{\mathrm{MCS}} 
    \label{eq:supp-tts-seconds}
\end{equation}
We obtain $\overline{\Delta t}_{\mathrm{MCS}}$ by dividing the measured execution time by the number of runs and the number of MCS. The reported timing represents the end-to-end optimization time for each method, conditional on a pre-generated proposal bank. For PT, the timing covers the complete optimization run. For TAPT, it includes all operations performed during the run (see Methods). Only offline generation of the proposal bank is excluded. Since this scaling study uses long-run MCMC configurations as the proposal source, no IsingFormer training or inference cost is involved.

We characterize problem size by the maximum product bit length $n$, as used in Fig.~\ref{fig:Figure TTS}. For equal-bit-length factors $A$ and $B$, the number of spins in the corresponding multiplier Ising model grows approximately quadratically with $n$. A fit to the spin counts of the constructed multiplier circuits gives
\begin{equation}
    N_{\mathrm{spin}}
    =
    \frac{3}{4}n^2+\frac{1}{2}n .
    \label{eq:supp-circuit-to-spins}
\end{equation}
Thus, increasing the product bit length also systematically increases the size of the underlying Ising optimization problem.

As shown in Fig.~\ref{fig:Figure TTS}, the median TTS is fit to
\begin{equation}
    \mathrm{TTS}(n)=A\exp(kn)
\end{equation}
where the scaling exponent $k$ is obtained from a linear least-squares fit of $\log[\mathrm{TTS}(n)]$ versus problem size $n$. The fitted exponents are $k_{\mathrm{PT}}=0.846$ and $k_{\mathrm{TAPT}}=0.564$. Thus, the fitted PT exponent is approximately $1.50$ times the TAPT exponent, corresponding to an approximately $33\%$ reduction in the fitted exponent.

\subsection{TRAINED ISINGFORMERS AND PROPOSAL GENERATION}
\label{supp: IsingFormersTraining}

The IsingFormer is trained using cross-entropy loss on configurations generated by long-run MCMC. The IsingFormer architecture is adjusted to the problem class. For the 2D ferromagnetic Ising model, we use a compact decoder-only Transformer with $d_{\mathrm{model}}=64$, $h=2$ attention heads, $L=2$ Transformer layers, and an FFN dimension of $2d_{\mathrm{model}}=128$. For the 3D spin-glass and factorization problems, we use a larger model with $d_{\mathrm{model}}=256$, $h=4$ attention heads, $L=5$ Transformer layers, and an FFN dimension of $4d_{\mathrm{model}}=1024$. In all cases, the model uses causal masked self-attention, sinusoidal positional encodings, and a learnable inverse-temperature embedding $e_{\beta}$, and is trained autoregressively via next-token prediction on long-run MCMC configurations generated at multiple $\beta$ values.

For temperature conditioning, the scalar inverse temperature $\beta$ is mapped to the model dimension through a learned linear projection, $e_\beta=W_\beta\beta+b_\beta$. The resulting embedding is added to the token embedding at every sequence position before positional encoding and Transformer processing. Because $\beta$ is represented as a continuous scalar rather than through a discrete lookup table, the same model can be evaluated directly at intermediate $\beta$ values not used during training.

A fixed spin ordering is used for each problem. For the 2D Ising model, we use a zigzag ordering of the lattice. For factorization, the spins are ordered as
\[
\{\text{product } C\}\{\text{carry bits}\}\{\text{digital multiplier}\}
\{\text{factor }B\}\{\text{factor }A\}
\]
Placing the product and carry bits first allows them to be provided as conditioning tokens during factorization inference. For the 3D spin-glass instances, no simple geometric ordering is imposed, and the fixed spin indexing of each instance is used consistently.

In the 2D Ising and 3D spin-glass experiments, the training datasets consist of unconditioned long-run MCMC configurations at different $\beta$ values. No clamping is applied, and these configurations are intended to approximate the corresponding Boltzmann distributions. For each training $\beta$, $10^{5}$ configurations are used, such that the sample counts reported in Table~\ref{tab:Trained_IsingFormer} correspond to the number of training $\beta$ values multiplied by $10^{5}$. For the 3D spin-glass experiment, a separate IsingFormer is trained for each of the 20 instances.

The nominal $L=10$ source graph contains 999 active spins rather than $10^3$ because one site is absent owing to the source hardware embedding~\citep{king2023quantum,chowdhury2025pushing}.

For semiprime factorization, a single IsingFormer is trained using long-run MCMC configurations from the universal free-running multiplier circuit, with no product instance imposed during training. During inference, the same trained IsingFormer is reused across semiprime instances by conditioning on the desired product bits $C$, together with the carry-in bits, which are fixed to zero. The remaining spins are then generated autoregressively. This allows the training cost to be amortized across different semiprime instances of the same multiplier size.

\begin{table}[t]
 \centering
 \caption{Trained IsingFormers}
 \label{tab:Trained_IsingFormer}
 \setlength{\tabcolsep}{3pt}
 \renewcommand{\arraystretch}{0.95}
 \scriptsize
 \begin{adjustbox}{width=\columnwidth}
 \begin{tabular}{lccccccc}
 \toprule
 \textbf{Instance Type} 
 & \textbf{Size} 
 & \makecell{\textbf{Number of}\\\textbf{Samples/Instance}} 
 & \makecell{\textbf{Number}\\\textbf{of Instances}} 
 & \textbf{$\beta$-range} 
 & \makecell{\textbf{Burn-in}\\\textbf{MCS}}  
 & \makecell{\textbf{Minutes/Epochs}\\\textbf{@ Batch}} 
 & \makecell{\textbf{Max}\\\textbf{Epochs}} \\
 \midrule

 Ferromagnetic
   & $50{\times}50\;(2500)$ 
   & $14\times10^{5}$ 
   & 1   
   & $[0.125,1.125]$ 
   & $10^{4}$ 
   & \texttt{40 @ 64} 
   & 100 \\

 \addlinespace[2pt]

 3D spin glass
   & $L=10\;(999)$ 
   & $19\times10^{5}$  
   & 20
   & $[0.5,2.204]$ 
   & $10^{4}$ 
   & \texttt{30 @ 512} 
   & 30 \\

 \addlinespace[2pt]

 Factorization
   & $20$-bit $(310)$
   & $5\times10^{5}$ 
   & 1 
   & $[0.5,1.250]$  
   & $10^{5}$ 
   & \texttt{20 @ 128}  
   & 40 \\

 \bottomrule
 \end{tabular}
 \end{adjustbox}
\end{table}

Training is performed on an \textit{NVIDIA RTX 6000 Ada} GPU. We use the AdamW optimizer with an initial learning rate of $10^{-4}$ and a weight decay of $3\times10^{-5}$. The learning rate is linearly warmed up over the first three epochs and subsequently follows a cosine-annealing schedule. We use a 90/10 training--validation split and apply gradient clipping with a maximum norm of $1.0$. The IsingFormer is trained for a fixed maximum number of epochs, as reported in Table~\ref{tab:Trained_IsingFormer}, and the final model is selected based on the checkpoint with the lowest validation loss. In terms of wall-clock time, the $50 \times 50$ 2D Ising model required roughly 3 days of training ($\sim$4000 minutes), while each $L=10$ 3D spin-glass IsingFormer required roughly 15 hours ($\sim$900 minutes). The corresponding training details for the factorization model are reported in Table~\ref{tab:Trained_IsingFormer}.

\begin{table*}[h]
\centering
\caption{Long-run MCMC proposal generation.}
\label{tab:MCMC_proposal_generation}
\setlength{\tabcolsep}{10pt}
\renewcommand{\arraystretch}{1.0}
\small
\begin{tabular}{lcccc}
\toprule
\textbf{Instance Type}
& \textbf{Size}
& \makecell{\textbf{Burn-in}\\\textbf{MCS}}
& \makecell{\textbf{Number of}\\\textbf{Generated Samples}}
& \textbf{Total Time (s)} \\
\midrule

3D spin glass
& $L=10\;(999)$
& $10^4$
& $10^5$
& $16.23$ \\

Factorization
& $16$-bit $(200)$
& $10^{5}$
& $10^{5}$
& $38.24$ \\

Factorization
& $18$-bit $(252)$
& $10^{5}$
& $10^{5}$
& $47.96$ \\

Factorization
& $20$-bit $(310)$
& $10^{5}$
& $10^{5}$
& $57.56$ \\

Factorization
& $22$-bit $(374)$
& $10^{5}$
& $10^{5}$
& $69.44$ \\

Factorization
& $24$-bit $(444)$
& $10^{5}$
& $10^{5}$
& $85.90$ \\

\bottomrule
\end{tabular}
\end{table*}

For experiments using long-run MCMC as the proposal generator, proposal samples are generally generated offline at the corresponding $\beta$ values and subsequently used within TAPT in place of IsingFormer proposals. In the 3D spin-glass study, these pre-generated MCMC proposals are used for comparison with IsingFormer and for the proposal-range experiments. The contextual proposals in Fig.~\ref{fig:replica,context}(b) are instead generated online under the current-replica clamping described above. For the factorization scaling study, MCMC-generated proposals are used to isolate the TAPT framework from the performance of a particular learned generator.

For the pre-generated proposal banks, each proposal sample is generated independently from a separately initialized MCMC chain. Specifically, for each sample, a new chain is randomly initialized and evolved for the burn-in period reported in Table~\ref{tab:MCMC_proposal_generation}, after which a single configuration is saved. For the 3D spin-glass experiments, we use a burn-in period of $10^{4}$ MCS for each generated sample, whereas for factorization we use $10^{5}$ MCS. The reported generation time corresponds to producing the complete $10^{5}$-sample proposal bank for a single $\beta$ value and a single problem instance. The same pre-generated proposal bank for a given problem instance and $\beta$ value is reused across all 400 independent TAPT runs, with each run drawing proposals from this common bank. All in-run operations are included in the end-to-end optimization timing.

\end{document}